\documentclass[10pt,twocolumn,letterpaper]{article}

\usepackage{wacv}
\usepackage{times}
\usepackage{epsfig}
\usepackage{graphicx}
\usepackage{amsmath}
\usepackage{amssymb}
\usepackage{subfigure}
\usepackage{xspace}

\usepackage{microtype}

\usepackage{xcolor}
\usepackage{textcomp}
\usepackage{color}
\usepackage{colortbl}
\usepackage{url}
\usepackage{gensymb}
\usepackage{multirow}

\graphicspath{{./graphics/}}
\usepackage{animate}
\usepackage{tabularx}
\usepackage{booktabs}
\usepackage{caption}
\usepackage{tikz}
\usepackage{pgfplots}
\usepackage{pgfplotstable}
\usepackage[skins]{tcolorbox}
\usepackage{standalone}
\usepackage{bm}
\usepackage{mleftright}
\usepackage{nicefrac}
\usepackage{makecell}
\usepackage{textcomp}
\usepackage[normalem]{ulem}
\usepackage[shortcuts]{extdash}
\usepackage[outline]{contour}
\newlength{\itemwidth}

\newcolumntype{Y}{>{\centering\arraybackslash}X}
\newcolumntype{P}[1]{>{\centering\arraybackslash}p{#1}}

\usetikzlibrary{calc}
\usetikzlibrary{tikzmark}
\usetikzlibrary{spy}
\usetikzlibrary{shapes.misc}
\pgfplotsset{compat=newest}

\definecolor{EE7F0E}{RGB}{238,127,14}
\definecolor{619D47}{RGB}{97,157,71}
\definecolor{3787CF}{RGB}{55,135,207}
\definecolor{DBDC4A}{RGB}{219,220,74}
\definecolor{3787CF}{RGB}{55,135,207}

\pgfplotscreateplotcyclelist{custompalette}{
    {EE7F0E!90!black,fill=EE7F0E},
    {619D47!90!black,fill=619D47},
    {3787CF!90!black,fill=3787CF},
    {DBDC4A!90!black,fill=DBDC4A}
}
\pgfplotscreateplotcyclelist{anothercustompalette}{
    {EE7F0E!90!black,line width=1.5pt},
    {619D47!90!black,line width=1.5pt},
    {3787CF!90!black,line width=1.5pt},
    {DBDC4A!90!black,line width=1.5pt}
}

\definecolor{3399FF}{RGB}{51, 153, 255}
\definecolor{FF9933}{RGB}{255, 153, 51}

\pgfkeys{
    cubepgf/.is family,
    cubepgf,
    x/.estore in = \cubex,
    y/.estore in = \cubey,
    width/.estore in = \cubewidth,
    height/.estore in = \cubeheight,
    depth/.estore in = \cubedepth,
    thickness/.estore in = \cubethickness,
    draw/.estore in = \cubedraw,
    fill/.estore in = \cubefill,
}

\usepackage{stmaryrd}
\usepackage{trimclip}

\makeatletter
\DeclareRobustCommand{\shortto}{%
  \mathrel{\mathpalette\short@to\relax}%
}

\DeclareRobustCommand{\veryshortto}{%
  \mathrel{\mathpalette\veryshort@to\relax}%
}

\newcommand{\short@to}[2]{%
  \mkern2mu
  \clipbox{{.3\width} 0 0 0}{$\m@th#1\vphantom{+}{\shortrightarrow}$}%
  }

\newcommand{\veryshort@to}[2]{%
  \mkern2mu
  \clipbox{{.2\width} 0 0 0}{$\m@th#1\vphantom{+}{\shortrightarrow}$}%
  }
\makeatother

\tikzset{
  double arrow/.style args={#1 with #2 and #3}{
    -stealth, line width=#1, #2, postaction={
        draw, -stealth, line width=(#1)/2, shorten <= (#1)/4, shorten >= 2*(#1)/4, #3
    }
  }
}

\tikzset{
  double arrow bothdir/.style args={#1 with #2 and #3}{
    stealth-stealth, line width=#1, #2, postaction={
        draw, stealth-stealth, line width=(#1)/2, shorten <= 2*(#1)/4, shorten >= 2*(#1)/4, #3
    }
  }
}

\newcommand{\usolid}[1]{%
    \tikz[remember picture, baseline=(tosolid.base)]{
        \node[inner sep=0pt, outer sep=0pt] (tosolid) {#1};
    }%
    \tikz[remember picture, overlay]{
        \draw[] ([yshift=-1.5pt]tosolid.south west) -- ([yshift=-1.5pt]tosolid.south east);
    }%
}%

\newcommand{\udotted}[1]{%
    \tikz[remember picture, baseline=(todotted.base)]{
        \node[inner sep=0pt, outer sep=0pt] (todotted) {#1};
    }%
    \tikz[remember picture, overlay]{
        \draw[densely dotted, line width=0.5] ([yshift=-1.5pt]todotted.south west) -- ([yshift=-1.5pt]todotted.south east);
    }%
}%

\newcommand{\uarrow}[1]{%
    \tikz[remember picture, baseline=(toarrow.base)]{
        \node[inner sep=0pt, outer sep=0pt] (toarrow) {#1};
    }%
    \tikz[remember picture, overlay]{
        \draw[->] ([yshift=-1.5pt]toarrow.south west) -- ([yshift=-1.5pt]toarrow.south east);
    }%
}%

\wacvfinalcopy

\usepackage[pagebackref=true,breaklinks=true,colorlinks,bookmarks=false]{hyperref}

\begin{document}

\title{Revisiting Adaptive Convolutions for Video Frame Interpolation\vspace{0.1cm}}

\makeatletter
\g@addto@macro\@maketitle{
    \vspace*{-13pt}
    \begin{center}\centering
        \setlength{\tabcolsep}{0.05cm}
        \setlength{\itemwidth}{4.25cm}
        \hspace*{-\tabcolsep}\begin{tabular}{cccc}
                \begin{tikzpicture}
                    \definecolor{arrowcolor}{RGB}{238,127,14}
                    \node [anchor=south west, inner sep=0.0cm] (image) at (0,0) {
                        \includegraphics[width=\itemwidth, trim={8.5cm 1.5cm 3.5cm 2.5cm}, clip]{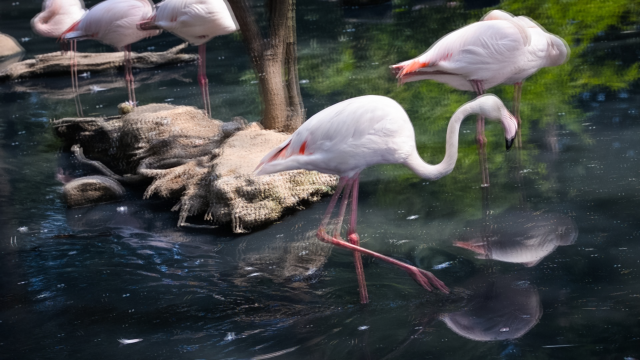}
                    };
                    \begin{scope}[x={(image.south east)},y={(image.north west)}]
                        \draw [double arrow=0.2cm with white and arrowcolor] (0.65,0.35) -- (0.45,0.42);
                    \end{scope}
                \end{tikzpicture}
            &
                \begin{tikzpicture}
                    \definecolor{arrowcolor}{RGB}{238,127,14}
                    \node [anchor=south west, inner sep=0.0cm] (image) at (0,0) {
                        \includegraphics[width=\itemwidth, trim={8.5cm 1.5cm 3.5cm 2.5cm}, clip]{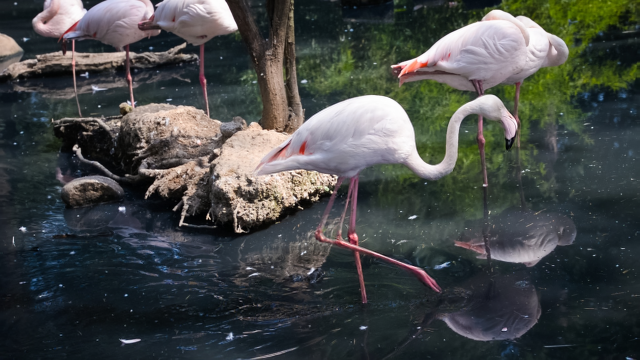}
                    };
                    \begin{scope}[x={(image.south east)},y={(image.north west)}]
                        \draw [double arrow=0.2cm with white and arrowcolor] (0.65,0.35) -- (0.45,0.42);
                    \end{scope}
                \end{tikzpicture}
            &
                \begin{tikzpicture}
                    \definecolor{arrowcolor}{RGB}{97,157,71}
                    \node [anchor=south west, inner sep=0.0cm] (image) at (0,0) {
                        \includegraphics[width=\itemwidth, trim={8.5cm 1.5cm 3.5cm 2.5cm}, clip]{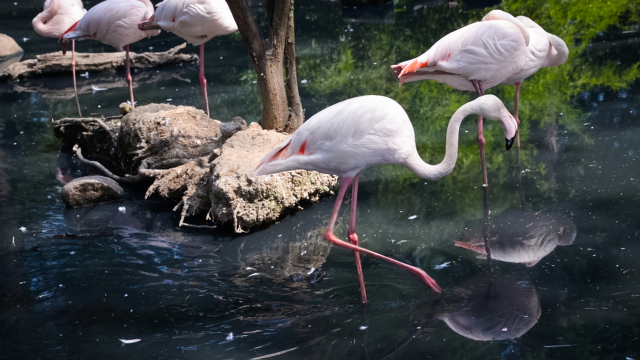}
                    };
                    \begin{scope}[x={(image.south east)},y={(image.north west)}]
                        \draw [double arrow=0.2cm with white and arrowcolor] (0.65,0.35) -- (0.45,0.42);
                    \end{scope}
                \end{tikzpicture}
            &
                \begin{tikzpicture}
                    \definecolor{arrowcolor}{RGB}{97,157,71}
                    \node [anchor=south west, inner sep=0.0cm] (image) at (0,0) {
                        \includegraphics[width=\itemwidth, trim={8.5cm 1.5cm 3.5cm 2.5cm}, clip]{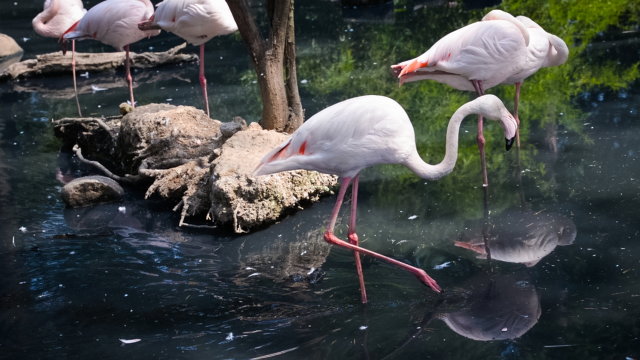}
                    };
                    \begin{scope}[x={(image.south east)},y={(image.north west)}]
                        \draw [double arrow=0.2cm with white and arrowcolor] (0.65,0.35) -- (0.45,0.42);
                    \end{scope}
                \end{tikzpicture}
            \\
                \footnotesize (a) overlayed input frames
            &
                \footnotesize (b) original SepConv~\cite{Niklaus_ICCV_2017}
            &
                \footnotesize (c) our improved SepConv\raisebox{0.0ex}{\footnotesize++}
            &
                \footnotesize (d) current state of the art \cite{Niklaus_CVPR_2020}
            \\
        \end{tabular}\vspace{-0.2cm}
    	\captionof{figure}{Frame interpolation example where the leg of the flamingo is difficult to handle. Our techniques to improve the original SepConv~(b) enable us to synthesize results~(c) that are comparable to current state-of-the-art approaches~(d).}\vspace{0.3cm}
    	\label{fig:teaser}
    \end{center}
}
\makeatother

\author{
Simon Niklaus\\
{\small Adobe Research}
\and
Long Mai\\
{\small Adobe Research}
\and
Oliver Wang\\
{\small Adobe Research}
}

\maketitle
\thispagestyle{empty}

\begin{abstract}

    Video frame interpolation, the synthesis of novel views in time, is an increasingly popular research direction with many new papers further advancing the state of the art. But as each new method comes with a host of variables that affect the interpolation quality, it can be hard to tell what is actually important for this task. In this work, we show, somewhat surprisingly, that it is possible to achieve near state-of-the-art results with an older, simpler approach, namely adaptive separable convolutions, by a subtle set of low level improvements. In doing so, we propose a number of intuitive but effective techniques to improve the frame interpolation quality, which also have the potential to other related applications of adaptive convolutions such as burst image denoising, joint image filtering, or video prediction.

\end{abstract}

\vspace*{-0.3cm}
\section{Introduction}
\label{sec:intro}
Video frame interpolation, the synthesis of intermediate frames between existing frames of a video, is an important technique with applications in frame-rate conversion~\cite{Meyer_CVPR_2015}, video editing~\cite{Meyer_BMVC_2018}, novel view interpolation~\cite{Kalantari_TOG_2016}, video compression~\cite{Wu_ECCV_2018}, and motion blur synthesis~\cite{Brooks_CVPR_2019}. While the performance of video frame interpolation approaches has seen steady improvements, research efforts have become increasingly complex. For example, DAIN~\cite{Bao_CVPR_2019} combines optical flow estimation~\cite{Sun_CVPR_2018}, single image depth estimation~\cite{Li_CVPR_2018}, context-aware image synthesis~\cite{Niklaus_CVPR_2018}, and adaptive convolutions~\cite{Niklaus_CVPR_2017}. However, we show that somewhat surprisingly, it is possible to achieve near state-of-art results with an older, simpler approach by carefully optimizing its individual parts. Specifically, we revisit the idea of using adaptive separable convolutions~\cite{Niklaus_ICCV_2017} and augment it with a set of intuitive improvements. This optimized SepConv\raisebox{0.2ex}{\footnotesize++} ranks second among all published methods in the Middlebury benchmark~\cite{Baker_IJCV_2011}.

\begin{figure*}\centering\vspace*{-0.2cm}
    \setlength{\tabcolsep}{0.05cm}
    \setlength{\itemwidth}{3.41cm}
    \hspace*{-\tabcolsep}\begin{tabular}{ccc}
            \includegraphics[]{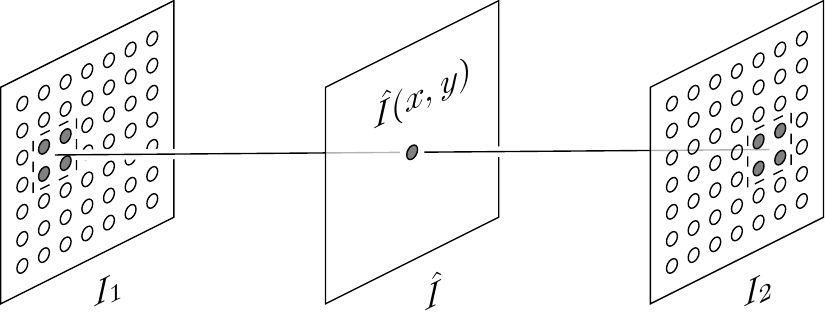}
        &
            \hspace{0.5cm}
        &
            \includegraphics[]{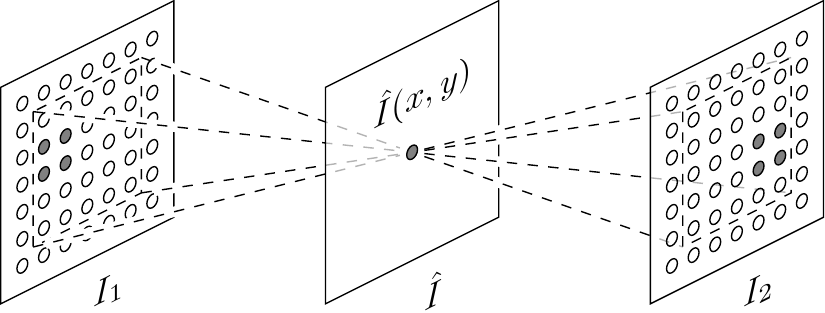}
        \\
            \footnotesize (a) flow-based interpolation with bilinearly-weighted sampling
        &
            
        &
            \footnotesize (b) kernel-based interpolation with spatially-varying kernels
        \\
    \end{tabular}\vspace{-0.2cm}
	\caption{Illustration of two prevalent video frame interpolation paradigms. Flow-based techniques first estimate the per-pixel motion between two frames and then compensate for it by warping the pixels according the estimated motion (a). This makes it possible to interpolate frames at an arbitrary time but one needs to account for inaccurate motion estimates and handle occlusions where optical flow is undefined. In comparison, frame interpolation via adaptive convolution jointly performs motion estimation and motion compensation in a single step by convolving input frames with spatially-varying kernels (b).}\vspace{-0.4cm}
	\label{fig:interparadigms}
\end{figure*}

The reason for choosing adaptive separable convolutions to show that an older frame interpolation method can be optimized to produce near state-of-the-art results are threefold. First, kernel-based video frame interpolation jointly performs motion estimation and motion compensation in a single step which makes for an elegant image formation model~\cite{Niklaus_CVPR_2017} (see Figure~\ref{fig:interparadigms} for more details on how this kernel-based interpolation differs from more-traditional flow-based interpolation). Second, adaptive separable convolutions are an efficient way to perform kernel-based interpolation~\cite{Niklaus_ICCV_2017}. In practice, filter kernels should be as large as possible to be able to account for large scene motion but this becomes prohibitively expensive with regular two-dimensional instead of two one-dimensional kernels. Third, adaptive convolutions have inspired and are part of many subsequent frame interpolation techniques~\cite{Bao_CVPR_2019, Bao_ARXIV_2018, Cheng_ARXIV_2020, Cheng_AAAI_2020, Lee_CVPR_2020}. As such, our findings on optimizing kernel-based video frame interpolation are directly applicable to the referenced approaches.

The idea of adaptive convolutions bears many names such as kernel prediction, dynamic filtering, basis prediction, or local attention. This technique has been proven effective in burst image denoising to align and merge multiple images~\cite{Marinc_ICIP_2019, Mildenhall_CVPR_2018, Xia_CVPR_2020}, in denoising Monte Carlo renderings by taking weighted averages of noisy neighborhoods~\cite{Bako_TOG_2017, Gharbi_TOG_2019, Vogels_TOG_2018}, in the modelling of a broad class of image transformations~\cite{Seitz_ICCV_2009}, in optical flow upsampling and joint image filtering~\cite{Kim_ARXIV_2019, Teed_ECCV_2020}, in video prediction where adaptive kernels can also model uncertainty~\cite{Finn_NIPS_2016, Jia_NIPS_2016, Reda_ECCV_2018, Xue_NIPS_2016}, in deblurring to model spatially-varying blur~\cite{Sim_OTHER_2019, Zhou_ICCV_2019}, or super-resolution where they can be used to merge multiple observations with sub-pixel accuracy~\cite{Cai_ICCV_2019, Jo_CVPR_2018}. While our paper focuses on improving adaptive separable convolutions for the purpose of frame interpolation, some of the improvements we introduce may be applicable in these related applications as well.

In summary, we revisit adaptive convolutions for frame interpolation and propose the following set of techniques that improve the method of SepConv~\cite{Niklaus_ICCV_2017} by a significant $1.76$~dB on the Middlebury benchmark examples~\cite{Baker_IJCV_2011} with publicly known ground truth (the relative improvement of each individual technique is shown in parenthesis).

\vspace{-0.0cm}
\begin{itemize}\itemsep-0.3em
    \item delayed padding ($+0.37$~dB)
    \item input normalization ($+0.30$~dB)
    \item network improvements ($+0.42$~dB)
    \item kernel normalization ($+0.52$~dB)
    \item contextual training ($+0.18$~dB)
    \item self-ensembling ($+0.18$~dB)
\end{itemize}
\vspace{-0.0cm}

These improvements allow our proposed SepConv\raisebox{0.2ex}{\footnotesize++} to quantitatively outperform all other frame interpolation approaches with the exception of SoftSplat~\cite{Niklaus_CVPR_2020} even though many of these methods are much more sophisticated.

\section{Related Work}
\label{sec:related}
With their work on adaptive convolutions for frame interpolation, Niklaus~\etal~\cite{Niklaus_CVPR_2017} proposed to perform joint motion estimation and motion compensation based on predicting spatially-varying kernels. This idea led to several interesting new developments such as the usage of adaptive separable convolutions~\cite{Niklaus_ICCV_2017}, adaptive warping layers that combine optical flow estimates and adaptive convolutions~\cite{Bao_CVPR_2019, Bao_ARXIV_2018}, additional per-coefficient offset vectors~\cite{Cheng_ARXIV_2020, Cheng_AAAI_2020}, spatially-varying deformable convolutions~\cite{Lee_CVPR_2020, Shi_ARXIV_2020}, or loss functions that leverage adaptive convolutions~\cite{Peleg_CVPR_2019}. Many of these efforts introduce novel ideas that enable smaller kernel sizes while simultaneously being able to compensate for arbitrarily-large motion. In this paper, we go back to the roots of kernel-based frame interpolation and revisit the idea of adaptive separable convolutions~\cite{Niklaus_ICCV_2017}, which strike a balance between simplicity and efficacy. By careful experimentation, we demonstrate several intuitive techniques that allow our proposed SepConv\raisebox{0.2ex}{\footnotesize++} to achieve near state-of-the-art results despite being based on an older approach.

\begin{figure*}\centering\vspace*{-0.2cm}
    \includegraphics[]{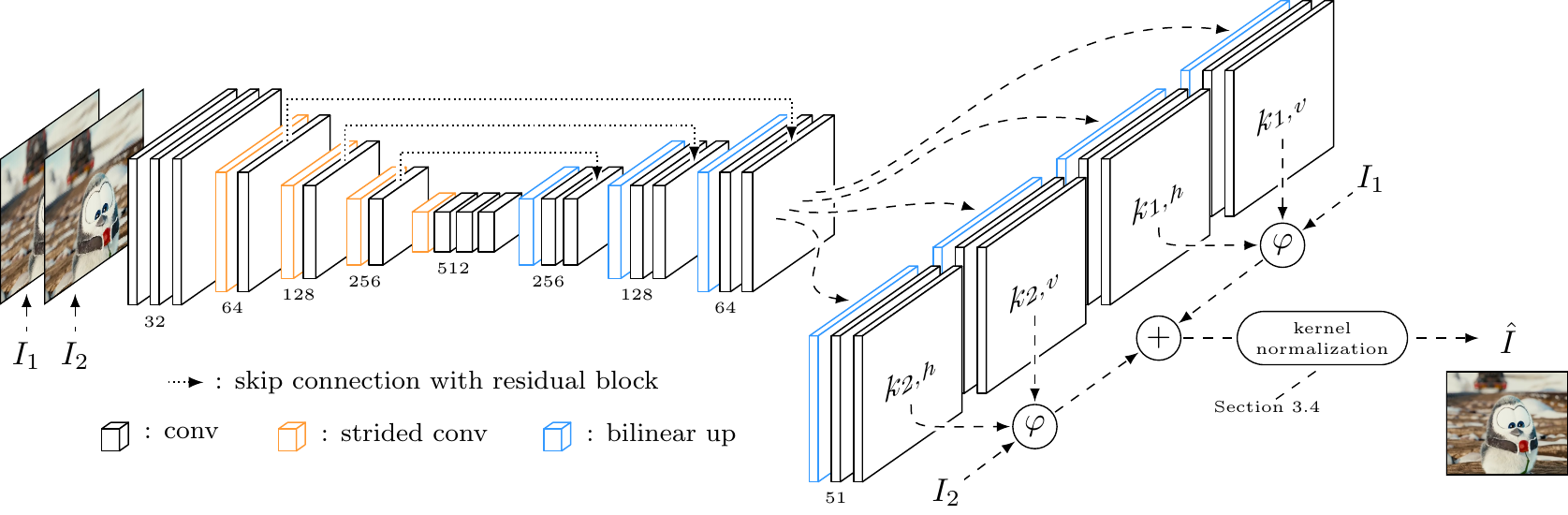}\vspace{-0.2cm}
	\caption{Overview of our frame interpolation framework where $\varphi$ denotes the adaptive separable convolution operator. We adopted the illustration style of the SepConv paper~\cite{Niklaus_ICCV_2017} to make it easier to compare our architecture with the original one.}\vspace{-0.4cm}
	\label{fig:architecture}
\end{figure*}

Aside from kernel-based interpolation, there are exciting research efforts that leverage explicit motion estimation in the form of optical flow for video frame interpolation. These efforts include estimating optical flow from the perspective of the frame that is ought to be synthesized~\cite{Liu_ICCV_2017}, not only warping the original frames but also their feature representations such that a synthesis network can predict better results from the additional context~\cite{Niklaus_CVPR_2018}, reconstructing optical flow representations from the perspective of the frame that is ought to be synthesized from a given inter-frame optical flow~\cite{Jiang_CVPR_2018}, fine-tuning the optical flow estimate for the given task at hand~\cite{Xue_IJCV_2019}, softmax splatting for differentiable forward warping in combination with feature pyramid synthesis~\cite{Niklaus_CVPR_2020}, leveraging multiple optical flow fields from the perspective of the frame that is ought to be synthesized~\cite{Park_ECCV_2020}, or utilizing a coarse-to-fine interpolation scheme~\cite{Amersfoort_ARXIV_2017, Zhang_ECCV_2020}. In comparison to kernel-based approaches, flow-based interpolation techniques have the advantage that their motion estimation component can be supervised on additional training data with ground truth optical flow. Despite this additional supervision, our proposed SepConv\raisebox{0.2ex}{\footnotesize++} still outperforms all flow-based methods with the exception of SoftSplat~\cite{Niklaus_CVPR_2020}.

Other approaches for frame interpolation that neither use adaptive convolutions nor optical flow include techniques based on phase~\cite{Meyer_CVPR_2018, Meyer_CVPR_2015} or approaches that directly synthesize the intermediate frame~\cite{Choi_AAAI_2020}. There are also interesting research efforts that perform frame interpolation in unison with a second video processing task like super-resolution~\cite{Kim_AAAI_2020, Xiang_CVPR_2020}, deblurring~\cite{Shen_CVPR_2020}, dequantization~\cite{Wang_CVPR_2019}, or with non-traditional acquisition setups~\cite{Lin_ECCV_2020, Paliwal_PAMI_2020, Wang_OTHER_2019}. Given that neural networks for frame interpolation are usually trained on off-the-shelf videos, it also seems natural to conduct research for test-time adaptation~\cite{Choi_CVPR_2020, Reda_ICCV_2019}. While most frame interpolation techniques only operate on two frames as input and hence assume linear motion, there are also interesting approaches that assume quadratic or cubic motion models~\cite{Chi_ECCV_2020, Liu_ARXIV_2020, Xu_NIPS_2019}. Our work is orthogonal to these ideas.

\section{Method}
\label{sec:method}
Given two consecutive frames $I_1$ and $I_2$ from a video, the frame interpolation task that we are targeting is the synthesis of the intermediate frame $\hat{I}$ that is temporally centered between the given input frames. To achieve this, we use the approach from Niklaus~\etal~\cite{Niklaus_ICCV_2017} that leverages adaptive separable convolutions by having a neural network $\phi$ predict a set of pixel-wise spatially-varying one-dimensional filter kernels $\langle K_{1,h}, K_{1,v}, K_{2,h}, K_{2,v} \rangle$ as follows.
\begin{equation}
    \langle K_{1,h}, K_{1,v}, K_{2,h}, K_{2,v} \rangle = \phi \left( I_1, I_2 \right)
\end{equation}
These spatially-varying kernels can then be used to process the input frames to yield $\hat{I}$ through an adaptive separable convolution operation $\varphi$. Specifically, $I_1$ is filtered with the separable filters $\langle K_{1,h}, K_{1,v} \rangle$ while $I_2$ is filtered with the separable filters $\langle K_{2,h}, K_{2,v} \rangle$ as follows.
\begin{equation}
    \hat{I} = \varphi \left( I_1, K_{1,h}, K_{1,v} \right) + \varphi \left( I_2, K_{2,h}, K_{2,v} \right)
\end{equation}
These spatially-varying kernels capture motion and re-sampling information, which makes for an effective image formation model for frame interpolation. To be able to account for large motion, the kernels should be as large as possible. However, with larger kernels it is more difficult to estimate all coefficients. We adopt the relatively moderate kernel size of $51$ pixels from the original SepConv~\cite{Niklaus_ICCV_2017}. We subsequently describe our proposed techniques to improve adaptive separable convolutions for frame interpolation.

\subsection{Delayed Padding}

As with all convolutions of non-singular size, the input needs to be padded if the output has to have the same resolution as the input. Specifically, the original SepConv~\cite{Niklaus_ICCV_2017} pads the input frames by $25$ pixels before estimating the adaptive kernel coefficients via a neural network $\phi$.
\begin{equation}
    \langle K_{1,h}, K_{1,v}, K_{2,h}, K_{2,v} \rangle = \phi \left( \text{\small pad}(I_1), \text{\small pad}(I_2) \right)
\end{equation}
In contrast, we propose not to pad the input images when they are given to $\phi$ but instead to pad them when the predicted kernels are applied to the input images as follows.
\begin{equation}
    \hat{I} = \varphi \left( \text{\small pad}(I_1), K_{1,h}, K_{1,v} \right) + \varphi \left( \text{\small pad}(I_2), K_{2,h}, K_{2,v} \right)
\end{equation}
This delayed padding has two positive effects. First, it improves the computational efficiency. Using an Nvidia V100, the original SepConv implementation takes $0.027$ seconds to interpolate a frame at a resolution of $512 \times 512$ pixels. In comparison, it takes $0.018$ seconds with the delayed padding. At a resolution of $1024 \times 1024$ pixels, it takes $0.083$ seconds with the original padding and $0.065$ seconds with our proposed delayed padding. Second, it improves the quality of the interpolated results since the neural network $\phi$ does not have to deal with large padded boundaries that are outside of the manifold of natural images (we use replication padding as in~\cite{Niklaus_ICCV_2017}). Specifically, delayed padding improves the interpolation results on the Middlebury benchmark examples~\cite{Baker_IJCV_2011} with publicly known ground truth by $0.37$~dB. Please see our ablation experiments in Section~\ref{sec:ablation} for more details.

\subsection{Input Normalization}

The contrast and brightness of the input frames should not affect the quality of the synthesized results. In other words, the network should be invariant to contrast and brightness. While it would be difficult to enforce such an invariance during training, it can easily be achieved by normalizing the input frames before feeding them to the network and denormalizing the synthesized result~\cite{Niklaus_CVPR_2018}. For image synthesis via adaptive convolutions, one can skip the denormalization step by applying the adaptive convolutions on the original input frames and only normalizing them when feeding them to the neural network that predicts the spatially-varying kernels.

To normalize the input frames, we shift and rescale their intensity values to have zero mean and unit standard deviation. There are multiple possibilities to do so, we have found normalizing the two images jointly while treating each color channel separately to work well. That is, for each color channel we compute the mean and standard deviation of $I_1$ and $I_2$ as if they were one image. This input normalization improves the interpolation quality on the Middlebury benchmark examples~\cite{Baker_IJCV_2011} with publicly known ground truth by $0.31$~dB. We also tried separately normalizing the input frames and computing singular instead of per-channel statistics but have found these approaches to be less effective.

One could make a similar argument about shift invariance since the quality of the synthesized interpolation results should not be affected by jointly translating the input frames. However, we have not been able to improve the interpolation quality by incorporating current techniques for improving the shift invariance of a neural network~\cite{Zhang_ICML_2019, Zou_ARXIV_2020}.

\begin{figure}\centering
    \hspace{-0.2cm}\includegraphics[]{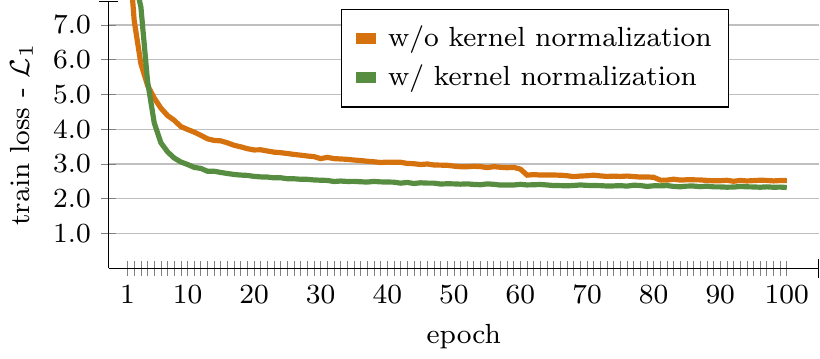}\vspace{-0.2cm}
	\caption{Comparing the training loss with and without our proposed kernel normalization, demonstrating that kernel normalization greatly improves the model convergence. Note that we halve the learning rate after $60$ and $80$ epochs.}\vspace{-0.4cm}
	\label{fig:kernorm}
\end{figure}

\subsection{Network Improvements}

Since the publciation of SepConv~\cite{Niklaus_ICCV_2017}, there has been great progress in deep learning architectures, and we experimented with incorporating these into an updated neural network architecture as shown in Figure~\ref{fig:architecture}. Specifically, we added residual blocks~\cite{He_CVPR_2016} to the skip connections that join the two halves of the U-Net, we changed the activation function to parametric rectified linear units~\cite{He_ICCV_2015}, we replaced the average pooling with strided convolutions, and we use a Kahan sum within the adaptive separable convolutions. Together, these changes lead to a $0.42$ dB improvement in terms of interpolation quality on the Middlebury benchmark examples~\cite{Baker_IJCV_2011} with publicly known ground truth.

These architectural changes do unfortunately not come for free. While it took $0.065$ seconds to synthesize a result at a resolution of $1024 \times 1024$ pixels using an Nvidia V100 with the original SepConv architecture, the new architecture takes $0.185$ seconds. This is surprising at first since our new architecture has fewer convolutions than the original one ($31$ versus $47$ convolutions) and our architecture is more parameter-efficient overall ($13.6$ million versus $21.7$ million parameters). However, our sub-networks that predict the adaptive kernel coefficients perform bilinear upsampling first whereas the original network performed bilinear upsampling last which leads to a significant increase in compute.

\begin{figure}\centering
    \setlength{\tabcolsep}{0.05cm}
    \setlength{\itemwidth}{4.11cm}
    \hspace*{-\tabcolsep}\begin{tabular}{cc}
            \begin{tikzpicture}
                \definecolor{arrowcolor}{RGB}{238,127,14}
                \node [anchor=south west, inner sep=0.0cm] (image) at (0,0) {
                    \includegraphics[width=\itemwidth, trim={3.0cm 7.3cm 9.0cm 0.0cm}, clip]{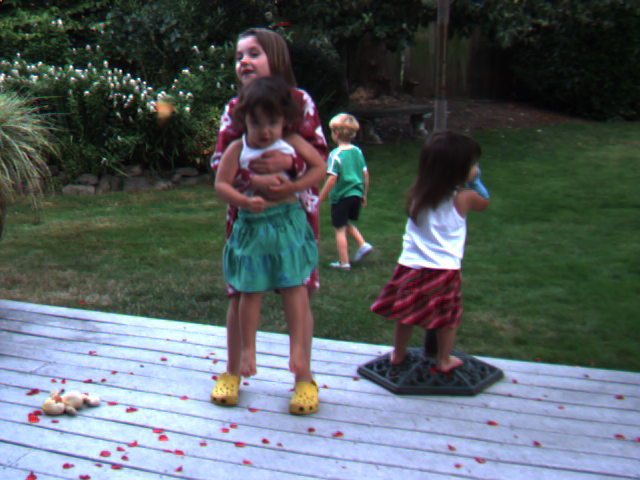}
                };
                \begin{scope}[x={(image.south east)},y={(image.north west)}]
                    \draw [double arrow=0.2cm with white and arrowcolor] (0.35,0.25) -- (0.28,0.5);
                \end{scope}
            \end{tikzpicture}
        &
            \begin{tikzpicture}
                \definecolor{arrowcolor}{RGB}{97,157,71}
                \node [anchor=south west, inner sep=0.0cm] (image) at (0,0) {
                    \includegraphics[width=\itemwidth, trim={3.0cm 7.3cm 9.0cm 0.0cm}, clip]{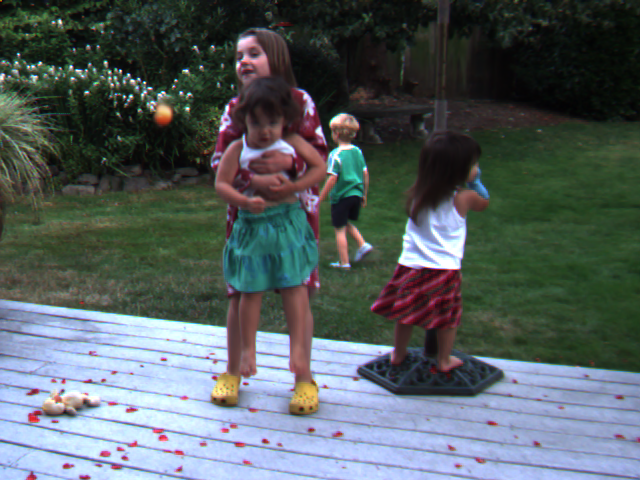}
                };
                \begin{scope}[x={(image.south east)},y={(image.north west)}]
                    \draw [double arrow=0.2cm with white and arrowcolor] (0.35,0.25) -- (0.28,0.5);
                \end{scope}
            \end{tikzpicture}
        \\
            \footnotesize (a) Ours - $\mathcal{L}_{\textit{Ctx}}$
        &
            \footnotesize (b) Ours - $\mathcal{L}_{\textit{Ctx}}$ - $8\hspace{-0.03cm}\times$
        \\
    \end{tabular}\vspace{-0.2cm}
	\caption{Comparing interpolation results without (a) and with (b) ensembling where eight independent predictions are combined (ensembling smooths uncertain estimates).}\vspace{-0.4cm}
	\label{fig:ensemble}
\end{figure}

\begin{figure*}\centering
    \setlength{\tabcolsep}{0.0cm}
    \renewcommand{\arraystretch}{1.2}
    \newcommand{\quantTit}[1]{\multicolumn{2}{c}{\scriptsize #1}}
    \newcommand{\quantSec}[1]{\scriptsize #1}
    \newcommand{\quantInd}[1]{\scriptsize #1}
    \newcommand{\quantVal}[1]{\scalebox{0.83}[1.0]{$ #1 $}}
    \newcommand{\quantFirst}[1]{\usolid{\scalebox{0.83}[1.0]{$ #1 $}}}
    \newcommand{\quantSecond}[1]{\udotted{\scalebox{0.83}[1.0]{$ #1 $}}}
    \footnotesize
    \begin{tabularx}{\textwidth}{@{\hspace{0.1cm}} X P{1.9cm} P{1.08cm} @{\hspace{-0.31cm}} P{1.85cm} P{1.08cm} @{\hspace{-0.31cm}} P{1.85cm} P{1.08cm} @{\hspace{-0.31cm}} P{1.85cm} P{1.08cm} @{\hspace{-0.31cm}} P{1.85cm} P{1.08cm} @{\hspace{-0.31cm}} P{1.85cm}}
        \toprule
            & & \quantTit{Middlebury} & \quantTit{Vimeo-90k} & \quantTit{UCF101 - DVF} & \quantTit{Xiph - 1K} & \quantTit{Xiph - 2K}
        \vspace{-0.1cm}\\
            & & \quantTit{Baker~\etal~\cite{Baker_IJCV_2011}} & \quantTit{Xue~\etal~\cite{Xue_IJCV_2019}} & \quantTit{Liu~\etal~\cite{Liu_ICCV_2017}} & \quantTit{(4K resized to 1K)} & \quantTit{(4K resized to 2K)}
        \\ \cmidrule(l{2pt}r{2pt}){3-4} \cmidrule(l{2pt}r{2pt}){5-6} \cmidrule(l{2pt}r{2pt}){7-8} \cmidrule(l{2pt}r{2pt}){9-10} \cmidrule(l{2pt}r{2pt}){11-12}
            & {\vspace{-0.29cm} \scriptsize training \linebreak dataset} & \quantSec{PSNR} \linebreak \quantInd{$\uparrow$} & {\vspace{-0.29cm} \scriptsize relative \linebreak improvement} & \quantSec{PSNR} \linebreak \quantInd{$\uparrow$} & {\vspace{-0.29cm} \scriptsize relative \linebreak improvement} & \quantSec{PSNR} \linebreak \quantInd{$\uparrow$} & {\vspace{-0.29cm} \scriptsize relative \linebreak improvement} & \quantSec{PSNR} \linebreak \quantInd{$\uparrow$} & {\vspace{-0.29cm} \scriptsize relative \linebreak improvement} & \quantSec{PSNR} \linebreak \quantInd{$\uparrow$} & {\vspace{-0.29cm} \scriptsize relative \linebreak improvement}
        \\ \midrule
original SepConv & proprietary & \quantVal{35.73} & \quantVal{-} & \quantVal{33.80} & \quantVal{-} & \quantVal{34.79} & \quantVal{-} & \quantVal{36.22} & \quantVal{-} & \quantVal{34.77} & \quantVal{-}
\\
reimplementation & Vimeo-90k & \quantVal{35.49} & \quantVal{-} & \quantVal{33.81} & \quantVal{-} & \quantVal{34.63} & \quantVal{-} & \quantVal{35.89} & \quantVal{-} & \quantVal{34.18} & \quantVal{-}
\\
+ delayed padding & --- \raisebox{-0.09cm}{''} --- & \quantVal{35.86} & \quantVal{\text{+ } 0.37 \text{ dB}} & \quantVal{34.31} & \quantVal{\text{+ } 0.50 \text{ dB}} & \quantVal{35.09} & \quantVal{\text{+ } 0.46 \text{ dB}} & \quantVal{36.00} & \quantVal{\text{+ } 0.11 \text{ dB}} & \quantVal{34.16} & \quantVal{\text{\scalebox{1.6}[1.0]{-} } 0.02 \text{ dB}}
\\
+ input normalization & --- \raisebox{-0.09cm}{''} --- & \quantVal{36.16} & \quantVal{\text{+ } 0.30 \text{ dB}} & \quantVal{34.50} & \quantVal{\text{+ } 0.19 \text{ dB}} & \quantVal{35.18} & \quantVal{\text{+ } 0.09 \text{ dB}} & \quantVal{36.06} & \quantVal{\text{+ } 0.06 \text{ dB}} & \quantVal{34.14} & \quantVal{\text{\scalebox{1.6}[1.0]{-} } 0.02 \text{ dB}}
\\
+ improved network & --- \raisebox{-0.09cm}{''} --- & \quantVal{36.58} & \quantVal{\text{+ } 0.42 \text{ dB}} & \quantVal{34.55} & \quantVal{\text{+ } 0.05 \text{ dB}} & \quantVal{35.19} & \quantVal{\text{+ } 0.01 \text{ dB}} & \quantVal{36.58} & \quantVal{\text{+ } 0.52 \text{ dB}} & \quantVal{34.76} & \quantVal{\text{+ } 0.62 \text{ dB}}
\\
+ normalized kernels & --- \raisebox{-0.09cm}{''} --- & \quantVal{37.10} & \quantVal{\text{+ } 0.52 \text{ dB}} & \quantVal{34.79} & \quantVal{\text{+ } 0.24 \text{ dB}} & \quantVal{35.22} & \quantVal{\text{+ } 0.03 \text{ dB}} & \quantVal{36.78} & \quantVal{\text{+ } 0.20 \text{ dB}} & \quantVal{34.77} & \quantVal{\text{+ } 0.01 \text{ dB}}
\\
+ contextual training & --- \raisebox{-0.09cm}{''} --- & \quantVal{37.28} & \quantVal{\text{+ } 0.18 \text{ dB}} & \quantVal{34.83} & \quantVal{\text{+ } 0.04 \text{ dB}} & \quantVal{35.24} & \quantVal{\text{+ } 0.02 \text{ dB}} & \quantVal{36.83} & \quantVal{\text{+ } 0.05 \text{ dB}} & \quantVal{34.84} & \quantVal{\text{+ } 0.07 \text{ dB}}
\\
+ self-ensembling & --- \raisebox{-0.09cm}{''} --- & \quantFirst{37.46} & \quantVal{\text{+ } 0.18 \text{ dB}} & \quantFirst{34.97} & \quantVal{\text{+ } 0.14 \text{ dB}} & \quantFirst{35.29} & \quantVal{\text{+ } 0.05 \text{ dB}} & \quantFirst{37.00} & \quantVal{\text{+ } 0.17 \text{ dB}} & \quantFirst{35.10} & \quantVal{\text{+ } 0.26 \text{ dB}}
        \\ \bottomrule
    \end{tabularx}\vspace{-0.2cm}
    \captionof{table}{Ablation experiments to quantitatively analyze the effects of our proposed techniques. In short, they each positively affect the interpolation quality across different dataset as long as the inter-frame motion does not exceed the kernel size.}\vspace{-0.4cm}
    \label{tbl:ablation}
\end{figure*}

\subsection{Kernel Normalization}

The initial paper on adaptive convolutions for video frame interpolation includes a softmax layer to normalize the kernels~\cite{Niklaus_CVPR_2017}, which is similar to but different from normalized convolutions~\cite{Knutsson_CVPR_1993} for addressing missing samples in the input signal. Such a kernel normalization is missing in the separable formulation since a softmax layer can no longer be used with this setup~\cite{Niklaus_ICCV_2017}. As a result, the neural network that predicts the kernel coefficients needs to take great care not to alter the apparent brightness of a synthesized pixel. We propose a simple normalization step that can be applied to any kernel-based image formation model. Specifically, we not only apply the adaptive separable convolution on the input but also on a singular mask. Afterwards, the filtered input can be divided by the filtered mask as follows such that denormalized kernel weights are compensated for.
\begin{equation}
    \hat{I} = \frac{ \varphi \left( I_1, K_{1,h}, K_{1,v} \right) + \varphi \left( I_2, K_{2,h}, K_{2,v} \right) }{ \varphi \left( {\bf 1}, K_{1,h}, K_{1,v} \right) + \varphi \left( {\bf 1}, K_{2,h}, K_{2,v} \right) }
\end{equation}
This simple kernel normalization step improves the quality of the synthesis results and greatly helps with the convergence of the model during training as shown in Figure~\ref{fig:kernorm}. With an improvement by $0.52$ dB on the Middlebury benchmark examples~\cite{Baker_IJCV_2011} with publicly known ground truth, the kernel normalization has the most significant impact on the quality of the synthesized results. Please see our ablation experiments in Section~\ref{sec:ablation} for more details.

\subsection{Contextual Training}

With adaptive convolutions for video frame interpolation, there is no constraint that forces the kernel prediction network to estimate coefficients that account for the true motion. Instead, the kernel prediction network may simply index pixels that have the desired color, which is similar to view synthesis by appearance flow~\cite{Zhou_ECCV_2016}. This may hurt the generalizability of the trained neural network though, which is why we force it to predict coefficients that agree with the true motion through a contextual loss. Specifically, we not only filter the input frames but also their context which have been obtained from an off-the-shelf network $\psi$ (we have found \texttt{relu1\_2} of a pre-trained VGG~\cite{Simonyan_ARXIV_2014} network and a trade-off weight $\alpha = 0.1$ to be effective). We then minimize not only the difference between the prediction and the ground truth in color space but also in the contextual space as follows (note that we omitted the previously introduced kernel normalization step in this definition for brevity).
\begin{equation}
    \mathcal{L}_{Ctx} = \left\| \langle \hat{I}, \alpha \cdot \hat{I}_\psi \rangle - \langle I_{gt}, \alpha \cdot \psi (I_{gt}) \rangle \right\| _1
\end{equation}
where
\begin{align} \begin{split}
    \hat{I} & = \varphi \left( I_1, K_{1,h}, K_{1,v} \right) + \varphi \left( I_2, K_{2,h}, K_{2,v} \right)
\end{split} \\ \begin{split}
    \hat{I}_\psi & = \varphi \left( \psi (I_1), K_{1,h}, K_{1,v} \right) + \varphi \left( \psi (I_2), K_{2,h}, K_{2,v} \right)
\end{split} \end{align}
Since each pixel in the contextual space not only describes the color of a single pixel but also encodes its local neighborhood, this loss effectively prevents the kernel prediction network from simply indexing pixels based on their color. Supervising the kernel prediction using this contextual loss yields an improvement of $0.18$ dB on the Middlebury benchmark examples~\cite{Baker_IJCV_2011} with publicly know ground truth.

Note that while this loss shares resemblance to a content loss~\cite{Johnson_ECCV_2016}, it is fundamentally different from it. A content loss applies a VGG feature loss directly on the synthesized result which would not prevent a kernel prediction network from estimating coefficients that mimic appearance flow. In contrast, we extract VGG features from the input frames before applying the adaptive separable convolution. In doing so, the kernel prediction network cannot just index a pixel with the same color as the ground truth as this may lead to significant differences in the VGG feature space.

\subsection{Self-ensembling}

In image classification~\cite{Chatfield_BMVC_2014} and super-resolution~\cite{Timofte_CVPR_2016}, a singular prediction is often enhanced by combining the predictions of multiple transformed versions of the same input. Such transforms include rotations, mirroring, or cropping. Not all image transforms are reversible though, but they have to be when wanting to combine predictions of pixel-wise tasks. Surprisingly, there is no study of the effect of such a self-ensembling approach in the context of frame interpolation. We hence propose to adopt this enhancement scheme for frame interpolation and conduct a large-scale study of its effect in Section~\ref{sec:ensembling}, demonstrating improvements across the board. In doing so, we consider taking the mean and taking the median of up to sixteen predictions with transforms based on reversing the input frames, flipping them, mirroring them, and applying rotations by ninety degrees.

Effectively, self-ensembling for video frame interpolation smooths predictions in areas where the kernel estimation is uncertain. As shown in Figure~\ref{fig:ensemble}, this can visually lead to a smooth result instead of one with visible artifacts. However, self-ensembling can computationally be prohibitively expensive as, for example, using eight predictions instead of just a single one will require eight times more compute.

\section{Experiments}
\label{sec:experiments}
\begin{figure*}\centering
    \setlength{\tabcolsep}{0.0cm}
    \renewcommand{\arraystretch}{1.2}
    \newcommand{\quantTit}[1]{\multicolumn{2}{c}{\scriptsize #1}}
    \newcommand{\quantSec}[1]{\scriptsize #1}
    \newcommand{\quantInd}[1]{\scriptsize #1}
    \newcommand{\quantVal}[1]{\scalebox{0.83}[1.0]{$ #1 $}}
    \newcommand{\quantFirst}[1]{\usolid{\scalebox{0.83}[1.0]{$ #1 $}}}
    \newcommand{\quantSecond}[1]{\udotted{\scalebox{0.83}[1.0]{$ #1 $}}}
    \footnotesize
    \begin{tabularx}{\textwidth}{@{\hspace{0.1cm}} X P{1.9cm} P{1.08cm} @{\hspace{-0.31cm}} P{1.85cm} P{1.08cm} @{\hspace{-0.31cm}} P{1.85cm} P{1.08cm} @{\hspace{-0.31cm}} P{1.85cm} P{1.08cm} @{\hspace{-0.31cm}} P{1.85cm} P{1.08cm} @{\hspace{-0.31cm}} P{1.85cm}}
        \toprule
            & & \quantTit{Middlebury} & \quantTit{Vimeo-90k} & \quantTit{UCF101 - DVF} & \quantTit{Xiph - 1K} & \quantTit{Xiph - 2K}
        \vspace{-0.1cm}\\
            & & \quantTit{Baker~\etal~\cite{Baker_IJCV_2011}} & \quantTit{Xue~\etal~\cite{Xue_IJCV_2019}} & \quantTit{Liu~\etal~\cite{Liu_ICCV_2017}} & \quantTit{(4K resized to 1K)} & \quantTit{(4K resized to 2K)}
        \\ \cmidrule(l{2pt}r{2pt}){3-4} \cmidrule(l{2pt}r{2pt}){5-6} \cmidrule(l{2pt}r{2pt}){7-8} \cmidrule(l{2pt}r{2pt}){9-10} \cmidrule(l{2pt}r{2pt}){11-12}
            & {\vspace{0.04cm} \scriptsize reduction} & \quantSec{PSNR} \linebreak \quantInd{$\uparrow$} & {\vspace{-0.29cm} \scriptsize relative \linebreak improvement} & \quantSec{PSNR} \linebreak \quantInd{$\uparrow$} & {\vspace{-0.29cm} \scriptsize relative \linebreak improvement} & \quantSec{PSNR} \linebreak \quantInd{$\uparrow$} & {\vspace{-0.29cm} \scriptsize relative \linebreak improvement} & \quantSec{PSNR} \linebreak \quantInd{$\uparrow$} & {\vspace{-0.29cm} \scriptsize relative \linebreak improvement} & \quantSec{PSNR} \linebreak \quantInd{$\uparrow$} & {\vspace{-0.29cm} \scriptsize relative \linebreak improvement}
        \\ \midrule
Ours - $\mathcal{L}_{\textit{Ctx}}$ & none & \quantVal{37.28} & \quantVal{-} & \quantVal{34.83} & \quantVal{-} & \quantVal{35.24} & \quantVal{-} & \quantVal{36.83} & \quantVal{-} & \quantVal{34.84} & \quantVal{-}
\\
Ours - $\mathcal{L}_{\textit{Ctx}}$ - $2\hspace{-0.03cm}\times$ & mean & \quantVal{37.41} & \quantVal{\text{+ } 0.13 \text{ dB}} & \quantVal{34.91} & \quantVal{\text{+ } 0.08 \text{ dB}} & \quantVal{35.26} & \quantVal{\text{+ } 0.02 \text{ dB}} & \quantVal{36.92} & \quantVal{\text{+ } 0.09 \text{ dB}} & \quantVal{35.01} & \quantVal{\text{+ } 0.17 \text{ dB}}
\\
Ours - $\mathcal{L}_{\textit{Ctx}}$ - $4\hspace{-0.03cm}\times$ & mean & \quantVal{37.45} & \quantVal{\text{+ } 0.04 \text{ dB}} & \quantVal{34.95} & \quantVal{\text{+ } 0.04 \text{ dB}} & \quantVal{35.28} & \quantVal{\text{+ } 0.02 \text{ dB}} & \quantVal{36.98} & \quantVal{\text{+ } 0.06 \text{ dB}} & \quantVal{35.09} & \quantVal{\text{+ } 0.08 \text{ dB}}
\\
Ours - $\mathcal{L}_{\textit{Ctx}}$ - $8\hspace{-0.03cm}\times$ & mean & \quantFirst{37.46} & \quantVal{\text{+ } 0.01 \text{ dB}} & \quantVal{34.97} & \quantVal{\text{+ } 0.02 \text{ dB}} & \quantVal{35.29} & \quantVal{\text{+ } 0.01 \text{ dB}} & \quantVal{37.00} & \quantVal{\text{+ } 0.02 \text{ dB}} & \quantVal{35.10} & \quantVal{\text{+ } 0.01 \text{ dB}}
\\
Ours - $\mathcal{L}_{\textit{Ctx}}$ - $16\hspace{-0.03cm}\times$ & mean & \quantFirst{37.46} & \quantVal{\text{+ } 0.00 \text{ dB}} & \quantFirst{34.99} & \quantVal{\text{+ } 0.02 \text{ dB}} & \quantFirst{35.30} & \quantVal{\text{+ } 0.01 \text{ dB}} & \quantFirst{37.02} & \quantVal{\text{+ } 0.02 \text{ dB}} & \quantFirst{35.13} & \quantVal{\text{+ } 0.03 \text{ dB}}
\\ \midrule
Ours - $\mathcal{L}_{\textit{Ctx}}$ & none & \quantVal{37.28} & \quantVal{-} & \quantVal{34.83} & \quantVal{-} & \quantVal{35.24} & \quantVal{-} & \quantVal{36.83} & \quantVal{-} & \quantVal{34.84} & \quantVal{-}
\\
Ours - $\mathcal{L}_{\textit{Ctx}}$ - $2\hspace{-0.03cm}\times$ & median & \quantVal{37.41} & \quantVal{\text{+ } 0.13 \text{ dB}} & \quantVal{34.91} & \quantVal{\text{+ } 0.08 \text{ dB}} & \quantVal{35.26} & \quantVal{\text{+ } 0.02 \text{ dB}} & \quantVal{36.92} & \quantVal{\text{+ } 0.09 \text{ dB}} & \quantVal{35.01} & \quantVal{\text{+ } 0.17 \text{ dB}}
\\
Ours - $\mathcal{L}_{\textit{Ctx}}$ - $4\hspace{-0.03cm}\times$ & median & \quantVal{37.44} & \quantVal{\text{+ } 0.03 \text{ dB}} & \quantVal{34.94} & \quantVal{\text{+ } 0.03 \text{ dB}} & \quantVal{35.27} & \quantVal{\text{+ } 0.01 \text{ dB}} & \quantVal{36.97} & \quantVal{\text{+ } 0.05 \text{ dB}} & \quantVal{35.07} & \quantVal{\text{+ } 0.06 \text{ dB}}
\\
Ours - $\mathcal{L}_{\textit{Ctx}}$ - $8\hspace{-0.03cm}\times$ & median & \quantFirst{37.47} & \quantVal{\text{+ } 0.03 \text{ dB}} & \quantVal{34.96} & \quantVal{\text{+ } 0.02 \text{ dB}} & \quantVal{35.28} & \quantVal{\text{+ } 0.01 \text{ dB}} & \quantVal{36.99} & \quantVal{\text{+ } 0.02 \text{ dB}} & \quantVal{35.09} & \quantVal{\text{+ } 0.02 \text{ dB}}
\\
Ours - $\mathcal{L}_{\textit{Ctx}}$ - $16\hspace{-0.03cm}\times$ & median & \quantFirst{37.47} & \quantVal{\text{+ } 0.00 \text{ dB}} & \quantFirst{34.98} & \quantVal{\text{+ } 0.02 \text{ dB}} & \quantFirst{35.29} & \quantVal{\text{+ } 0.01 \text{ dB}} & \quantFirst{37.01} & \quantVal{\text{+ } 0.02 \text{ dB}} & \quantFirst{35.13} & \quantVal{\text{+ } 0.04 \text{ dB}}
        \\ \bottomrule
    \end{tabularx}\vspace{-0.2cm}
    \captionof{table}{Effect of combining multiple independent predictions when taking their mean (top) and their median (bottom).}\vspace{-0.2cm}
    \label{tbl:ensemble}
\end{figure*}

\begin{figure*}\centering
    \setlength{\tabcolsep}{0.0cm}
    \renewcommand{\arraystretch}{1.2}
    \newcommand{\quantTit}[1]{\multicolumn{2}{c}{\scriptsize #1}}
    \newcommand{\quantSec}[1]{\scriptsize #1}
    \newcommand{\quantInd}[1]{\scriptsize #1}
    \newcommand{\quantVal}[1]{\scalebox{0.83}[1.0]{$ #1 $}}
    \newcommand{\quantFirst}[1]{\usolid{\scalebox{0.83}[1.0]{$ #1 $}}}
    \newcommand{\quantSecond}[1]{\udotted{\scalebox{0.83}[1.0]{$ #1 $}}}
    \footnotesize
    \begin{tabularx}{\textwidth}{@{\hspace{0.1cm}} X P{1.9cm} P{1.465cm} @{\hspace{-0.31cm}} P{1.465cm} P{1.465cm} @{\hspace{-0.31cm}} P{1.465cm} P{1.465cm} @{\hspace{-0.31cm}} P{1.465cm} P{1.465cm} @{\hspace{-0.31cm}} P{1.465cm} P{1.465cm} @{\hspace{-0.31cm}} P{1.465cm}}
        \toprule
            & & \quantTit{Middlebury} & \quantTit{Vimeo-90k} & \quantTit{UCF101 - DVF} & \quantTit{Xiph - 1K} & \quantTit{Xiph - 2K}
        \vspace{-0.1cm}\\
            & & \quantTit{Baker~\etal~\cite{Baker_IJCV_2011}} & \quantTit{Xue~\etal~\cite{Xue_IJCV_2019}} & \quantTit{Liu~\etal~\cite{Liu_ICCV_2017}} & \quantTit{(4K resized to 1K)} & \quantTit{(4K resized to 2K)}
        \\ \cmidrule(l{2pt}r{2pt}){3-4} \cmidrule(l{2pt}r{2pt}){5-6} \cmidrule(l{2pt}r{2pt}){7-8} \cmidrule(l{2pt}r{2pt}){9-10} \cmidrule(l{2pt}r{2pt}){11-12}
            & {\vspace{0.04cm} \scriptsize venue} & {\vspace{-0.29cm} \scriptsize without \linebreak ensemble} & {\vspace{-0.29cm} \scriptsize with \linebreak ensemble} & {\vspace{-0.29cm} \scriptsize without \linebreak ensemble} & {\vspace{-0.29cm} \scriptsize with \linebreak ensemble} & {\vspace{-0.29cm} \scriptsize without \linebreak ensemble} & {\vspace{-0.29cm} \scriptsize with \linebreak ensemble} & {\vspace{-0.29cm} \scriptsize without \linebreak ensemble} & {\vspace{-0.29cm} \scriptsize with \linebreak ensemble} & {\vspace{-0.29cm} \scriptsize without \linebreak ensemble} & {\vspace{-0.29cm} \scriptsize with \linebreak ensemble}
        \\ \midrule
SepConv - $\mathcal{L}_1$ & ICCV 2017 & \quantVal{35.73}\makebox[0cm]{\raisebox{0.065cm}{\hspace{0.6cm}\scalebox{0.6}[0.9]{\tiny\uarrow{\thinspace + $0.41$ \thinspace\thinspace}}}} & \quantVal{36.14} & \quantVal{33.80}\makebox[0cm]{\raisebox{0.065cm}{\hspace{0.6cm}\scalebox{0.6}[0.9]{\tiny\uarrow{\thinspace + $0.24$ \thinspace\thinspace}}}} & \quantVal{34.04} & \quantVal{34.79}\makebox[0cm]{\raisebox{0.065cm}{\hspace{0.6cm}\scalebox{0.6}[0.9]{\tiny\uarrow{\thinspace + $0.14$ \thinspace\thinspace}}}} & \quantVal{34.93} & \quantVal{36.22}\makebox[0cm]{\raisebox{0.065cm}{\hspace{0.6cm}\scalebox{0.6}[0.9]{\tiny\uarrow{\thinspace + $0.24$ \thinspace\thinspace}}}} & \quantVal{36.46} & \quantVal{34.77}\makebox[0cm]{\raisebox{0.065cm}{\hspace{0.6cm}\scalebox{0.6}[0.9]{\tiny\uarrow{\thinspace + $0.39$ \thinspace\thinspace}}}} & \quantVal{35.16}
\\
CtxSyn - $\mathcal{L}_{\textit{Lap}}$ & CVPR 2018 & \quantVal{36.93}\makebox[0cm]{\raisebox{0.065cm}{\hspace{0.6cm}\scalebox{0.6}[0.9]{\tiny\uarrow{\thinspace + $0.39$ \thinspace\thinspace}}}} & \quantVal{37.32} & \quantVal{34.39}\makebox[0cm]{\raisebox{0.065cm}{\hspace{0.6cm}\scalebox{0.6}[0.9]{\tiny\uarrow{\thinspace + $0.31$ \thinspace\thinspace}}}} & \quantVal{34.70} & \quantVal{34.62}\makebox[0cm]{\raisebox{0.065cm}{\hspace{0.6cm}\scalebox{0.6}[0.9]{\tiny\uarrow{\thinspace + $0.27$ \thinspace\thinspace}}}} & \quantVal{34.89} & \quantVal{36.87}\makebox[0cm]{\raisebox{0.065cm}{\hspace{0.6cm}\scalebox{0.6}[0.9]{\tiny\uarrow{\thinspace + $0.32$ \thinspace\thinspace}}}} & \quantVal{37.19} & \quantVal{35.72}\makebox[0cm]{\raisebox{0.065cm}{\hspace{0.6cm}\scalebox{0.6}[0.9]{\tiny\uarrow{\thinspace + $0.32$ \thinspace\thinspace}}}} & \quantVal{36.04}
\\
DAIN & CVPR 2019 & \quantVal{36.69}\makebox[0cm]{\raisebox{0.065cm}{\hspace{0.6cm}\scalebox{0.6}[0.9]{\tiny\uarrow{\thinspace + $0.35$ \thinspace\thinspace}}}} & \quantVal{37.04} & \quantVal{34.70}\makebox[0cm]{\raisebox{0.065cm}{\hspace{0.6cm}\scalebox{0.6}[0.9]{\tiny\uarrow{\thinspace + $0.25$ \thinspace\thinspace}}}} & \quantVal{34.95} & \quantVal{35.00}\makebox[0cm]{\raisebox{0.065cm}{\hspace{0.6cm}\scalebox{0.6}[0.9]{\tiny\uarrow{\thinspace + $0.13$ \thinspace\thinspace}}}} & \quantVal{35.13} & \quantVal{36.78}\makebox[0cm]{\raisebox{0.065cm}{\hspace{0.6cm}\scalebox{0.6}[0.9]{\tiny\uarrow{\thinspace + $0.26$ \thinspace\thinspace}}}} & \quantVal{37.04} & \quantVal{35.93}\makebox[0cm]{\raisebox{0.065cm}{\hspace{0.6cm}\scalebox{0.6}[0.9]{\tiny\uarrow{\thinspace + $0.26$ \thinspace\thinspace}}}} & \quantVal{36.19}
\\
CAIN & AAAI 2020 & \quantVal{35.11}\makebox[0cm]{\raisebox{0.065cm}{\hspace{0.6cm}\scalebox{0.6}[0.9]{\tiny\uarrow{\thinspace + $0.23$ \thinspace\thinspace}}}} & \quantVal{35.34} & \quantVal{34.65}\makebox[0cm]{\raisebox{0.065cm}{\hspace{0.6cm}\scalebox{0.6}[0.9]{\tiny\uarrow{\thinspace + $0.16$ \thinspace\thinspace}}}} & \quantVal{34.81} & \quantVal{34.98}\makebox[0cm]{\raisebox{0.065cm}{\hspace{0.6cm}\scalebox{0.6}[0.9]{\tiny\uarrow{\thinspace + $0.10$ \thinspace\thinspace}}}} & \quantVal{35.08} & \quantVal{36.21}\makebox[0cm]{\raisebox{0.065cm}{\hspace{0.6cm}\scalebox{0.6}[0.9]{\tiny\uarrow{\thinspace + $0.19$ \thinspace\thinspace}}}} & \quantVal{36.40} & \quantVal{35.18}\makebox[0cm]{\raisebox{0.065cm}{\hspace{0.6cm}\scalebox{0.6}[0.9]{\tiny\uarrow{\thinspace + $0.21$ \thinspace\thinspace}}}} & \quantVal{35.39}
\\
EDSC - $\mathcal{L}_C$ & arXiv 2020 & \quantVal{36.82}\makebox[0cm]{\raisebox{0.065cm}{\hspace{0.6cm}\scalebox{0.6}[0.9]{\tiny\uarrow{\thinspace + $0.44$ \thinspace\thinspace}}}} & \quantVal{37.26} & \quantVal{34.83}\makebox[0cm]{\raisebox{0.065cm}{\hspace{0.6cm}\scalebox{0.6}[0.9]{\tiny\uarrow{\thinspace + $0.27$ \thinspace\thinspace}}}} & \quantVal{35.10} & \quantVal{35.13}\makebox[0cm]{\raisebox{0.065cm}{\hspace{0.6cm}\scalebox{0.6}[0.9]{\tiny\uarrow{\thinspace + $0.09$ \thinspace\thinspace}}}} & \quantVal{35.22} & \quantVal{36.73}\makebox[0cm]{\raisebox{0.065cm}{\hspace{0.6cm}\scalebox{0.6}[0.9]{\tiny\uarrow{\thinspace + $0.31$ \thinspace\thinspace}}}} & \quantVal{37.04} & \quantVal{\text{OOM}}\makebox[0cm]{\raisebox{0.065cm}{\hspace{0.6cm}\scalebox{0.6}[0.9]{\tiny\uarrow{\thinspace + $0.00$ \thinspace\thinspace}}}} & \quantVal{\text{OOM}}
\\
AdaCoF & CVPR 2020 & \quantVal{35.72}\makebox[0cm]{\raisebox{0.065cm}{\hspace{0.6cm}\scalebox{0.6}[0.9]{\tiny\uarrow{\thinspace + $0.47$ \thinspace\thinspace}}}} & \quantVal{36.19} & \quantVal{34.35}\makebox[0cm]{\raisebox{0.065cm}{\hspace{0.6cm}\scalebox{0.6}[0.9]{\tiny\uarrow{\thinspace + $0.45$ \thinspace\thinspace}}}} & \quantVal{34.80} & \quantVal{35.16}\makebox[0cm]{\raisebox{0.065cm}{\hspace{0.6cm}\scalebox{0.6}[0.9]{\tiny\uarrow{\thinspace + $0.12$ \thinspace\thinspace}}}} & \quantVal{35.28} & \quantVal{36.26}\makebox[0cm]{\raisebox{0.065cm}{\hspace{0.6cm}\scalebox{0.6}[0.9]{\tiny\uarrow{\thinspace + $0.47$ \thinspace\thinspace}}}} & \quantVal{36.73} & \quantVal{34.82}\makebox[0cm]{\raisebox{0.065cm}{\hspace{0.6cm}\scalebox{0.6}[0.9]{\tiny\uarrow{\thinspace + $0.40$ \thinspace\thinspace}}}} & \quantVal{35.22}
\\
SoftSplat - $\mathcal{L}_{\textit{Lap}}$ & CVPR 2020 & \quantVal{38.42}\makebox[0cm]{\raisebox{0.065cm}{\hspace{0.6cm}\scalebox{0.6}[0.9]{\tiny\uarrow{\thinspace + $0.26$ \thinspace\thinspace}}}} & \quantVal{38.68} & \quantVal{36.10}\makebox[0cm]{\raisebox{0.065cm}{\hspace{0.6cm}\scalebox{0.6}[0.9]{\tiny\uarrow{\thinspace + $0.18$ \thinspace\thinspace}}}} & \quantVal{36.28} & \quantVal{35.39}\makebox[0cm]{\raisebox{0.065cm}{\hspace{0.6cm}\scalebox{0.6}[0.9]{\tiny\uarrow{\thinspace + $0.09$ \thinspace\thinspace}}}} & \quantVal{35.48} & \quantVal{37.96}\makebox[0cm]{\raisebox{0.065cm}{\hspace{0.6cm}\scalebox{0.6}[0.9]{\tiny\uarrow{\thinspace + $0.23$ \thinspace\thinspace}}}} & \quantVal{38.19} & \quantVal{36.63}\makebox[0cm]{\raisebox{0.065cm}{\hspace{0.6cm}\scalebox{0.6}[0.9]{\tiny\uarrow{\thinspace + $0.16$ \thinspace\thinspace}}}} & \quantVal{36.79}
\\
BMBC & ECCV 2020 & \quantVal{36.79}\makebox[0cm]{\raisebox{0.065cm}{\hspace{0.6cm}\scalebox{0.6}[0.9]{\tiny\uarrow{\thinspace + $0.12$ \thinspace\thinspace}}}} & \quantVal{36.91} & \quantVal{35.06}\makebox[0cm]{\raisebox{0.065cm}{\hspace{0.6cm}\scalebox{0.6}[0.9]{\tiny\uarrow{\thinspace + $0.15$ \thinspace\thinspace}}}} & \quantVal{35.21} & \quantVal{35.16}\makebox[0cm]{\raisebox{0.065cm}{\hspace{0.6cm}\scalebox{0.6}[0.9]{\tiny\uarrow{\thinspace + $0.08$ \thinspace\thinspace}}}} & \quantVal{35.24} & \quantVal{36.59}\makebox[0cm]{\raisebox{0.065cm}{\hspace{0.6cm}\scalebox{0.6}[0.9]{\tiny\uarrow{\thinspace + $0.37$ \thinspace\thinspace}}}} & \quantVal{36.96} & \quantVal{\text{OOM}}\makebox[0cm]{\raisebox{0.065cm}{\hspace{0.6cm}\scalebox{0.6}[0.9]{\tiny\uarrow{\thinspace + $0.00$ \thinspace\thinspace}}}} & \quantVal{\text{OOM}}
\\
Ours - $\mathcal{L}_{\textit{Ctx}}$ & N/A & \quantVal{37.28}\makebox[0cm]{\raisebox{0.065cm}{\hspace{0.6cm}\scalebox{0.6}[0.9]{\tiny\uarrow{\thinspace + $0.18$ \thinspace\thinspace}}}} & \quantVal{37.46} & \quantVal{34.83}\makebox[0cm]{\raisebox{0.065cm}{\hspace{0.6cm}\scalebox{0.6}[0.9]{\tiny\uarrow{\thinspace + $0.14$ \thinspace\thinspace}}}} & \quantVal{34.97} & \quantVal{35.24}\makebox[0cm]{\raisebox{0.065cm}{\hspace{0.6cm}\scalebox{0.6}[0.9]{\tiny\uarrow{\thinspace + $0.05$ \thinspace\thinspace}}}} & \quantVal{35.29} & \quantVal{36.83}\makebox[0cm]{\raisebox{0.065cm}{\hspace{0.6cm}\scalebox{0.6}[0.9]{\tiny\uarrow{\thinspace + $0.17$ \thinspace\thinspace}}}} & \quantVal{37.00} & \quantVal{34.84}\makebox[0cm]{\raisebox{0.065cm}{\hspace{0.6cm}\scalebox{0.6}[0.9]{\tiny\uarrow{\thinspace + $0.26$ \thinspace\thinspace}}}} & \quantVal{35.10}
        \\ \bottomrule
    \end{tabularx}\vspace{-0.2cm}
    \captionof{table}{Effect of combining the mean of eight independent predictions for several video frame interpolation methods.}\vspace{-0.4cm}
    \label{tbl:ensothers}
\end{figure*}

We subsequently evaluate our contributions by answering the following questions. What is the impact of each of our proposed techniques? What is the effect of self-ensembling for video frame interpolation? How does our SepConv\raisebox{0.2ex}{\footnotesize++} compare to the original SepConv? How does our SepConv\raisebox{0.2ex}{\footnotesize++} compare to other frame interpolation techniques?

\vspace{0.05in}
\noindent\textbf{Implementation.} We generally follow the implementation details of the original SepConv~\cite{Niklaus_ICCV_2017}. However, they were using a proprietary training dataset whereas our reimplementation is incorporating Vimeo-90k~\cite{Xue_IJCV_2019} instead. Furthermore, we simplified the augmentation pipeline and refrained from shifting the cropping windows in opposing directions.

\vspace{0.05in}
\noindent\textbf{Datasets.} We adopt the test set selection from~\cite{Niklaus_CVPR_2020} and conduct our evaluation on Vimeo-90k~\cite{Xue_IJCV_2019}, the Middlebury benchmark samples with publicly known ground truth~\cite{Baker_IJCV_2011}, the Liu~\etal~\cite{Liu_ICCV_2017} samples from UCF101~\cite{Soomro_ARXIV_2012}, and footage from Xiph\footnote{\url{https://media.xiph.org/video/derf}}. We do not adopt the ``4K'' cropped version of Xiph that~\cite{Niklaus_CVPR_2020} proposed though and instead focus on downscaled versions at a resolution of 1K as well as 2K.

\vspace{0.05in}
\noindent\textbf{Metrics.} We limit the evaluation herein to the PSNR metric since SSIM~\cite{Wang_TIP_2004} is subject to unexpected and unintuitive results~\cite{Nilsson_ARXIV_2020}. However, we provide equivalent tables with SSIM instead of PSNR in the supplementary material. These supplementary results support our claims and are generally aligned with PSNR in terms of relative improvements.

\begin{figure*}\centering
    \setlength{\tabcolsep}{0.05cm}
    \setlength{\itemwidth}{5.75cm}
    \hspace*{-\tabcolsep}\begin{tabular}{ccc}
            \begin{tikzpicture}
                \definecolor{arrowcolor}{RGB}{238,127,14}
                \node [anchor=south west, inner sep=0.0cm] (image) at (0,0) {
                    \includegraphics[width=\itemwidth, trim={0.0cm 0.0cm 0.0cm 0.0cm}, clip]{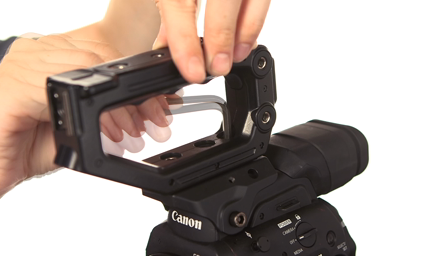}
                };
                \begin{scope}[x={(image.south east)},y={(image.north west)}]
                    \draw [double arrow=0.2cm with white and arrowcolor] (0.52,0.23) -- (0.36,0.38);
                \end{scope}
            \end{tikzpicture}
        &
            \begin{tikzpicture}
                \definecolor{arrowcolor}{RGB}{238,127,14}
                \node [anchor=south west, inner sep=0.0cm] (image) at (0,0) {
                    \includegraphics[width=\itemwidth, trim={0.0cm 0.0cm 0.0cm 0.0cm}, clip]{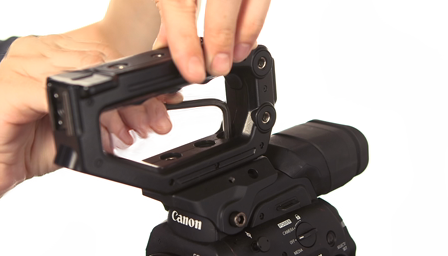}
                };
                \begin{scope}[x={(image.south east)},y={(image.north west)}]
                    \draw [double arrow=0.2cm with white and arrowcolor] (0.52,0.23) -- (0.36,0.38);
                \end{scope}
            \end{tikzpicture}
        &
            \begin{tikzpicture}
                \definecolor{arrowcolor}{RGB}{97,157,71}
                \node [anchor=south west, inner sep=0.0cm] (image) at (0,0) {
                    \includegraphics[width=\itemwidth, trim={0.0cm 0.0cm 0.0cm 0.0cm}, clip]{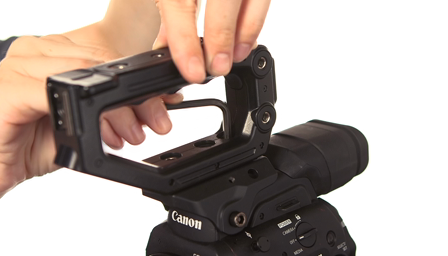}
                };
                \begin{scope}[x={(image.south east)},y={(image.north west)}]
                    \draw [double arrow=0.2cm with white and arrowcolor] (0.52,0.23) -- (0.36,0.38);
                \end{scope}
            \end{tikzpicture}
        \\
            \footnotesize (a) overlayed input frames
        &
            \footnotesize (b) original SepConv~\cite{Niklaus_ICCV_2017}
        &
            \footnotesize (c) our improved SepConv\raisebox{0.2ex}{\footnotesize++}
        \\
    \end{tabular}\vspace{-0.2cm}
    \caption{Qualitative comparison with SepConv. We purposefully only show a single example here for brevity and kindly refer to our supplementary video which shows this example as well as many more examples in a fully interpolated sequence.}\vspace{-0.2cm}
    \label{fig:qualitative}
\end{figure*}

\begin{figure*}\centering
    \setlength{\tabcolsep}{0.0cm}
    \renewcommand{\arraystretch}{1.2}
    \newcommand{\quantTit}[1]{\multicolumn{2}{c}{\scriptsize #1}}
    \newcommand{\quantSec}[1]{\scriptsize #1}
    \newcommand{\quantInd}[1]{\scriptsize #1}
    \newcommand{\quantVal}[1]{\scalebox{0.83}[1.0]{$ #1 $}}
    \newcommand{\quantFirst}[1]{\usolid{\scalebox{0.83}[1.0]{$ #1 $}}}
    \newcommand{\quantSecond}[1]{\udotted{\scalebox{0.83}[1.0]{$ #1 $}}}
    \footnotesize
    \begin{tabularx}{\textwidth}{@{\hspace{0.1cm}} X P{1.9cm} P{1.08cm} @{\hspace{-0.31cm}} P{1.85cm} P{1.08cm} @{\hspace{-0.31cm}} P{1.85cm} P{1.08cm} @{\hspace{-0.31cm}} P{1.85cm} P{1.08cm} @{\hspace{-0.31cm}} P{1.85cm} P{1.08cm} @{\hspace{-0.31cm}} P{1.85cm}}
        \toprule
            & & \quantTit{Middlebury} & \quantTit{Vimeo-90k} & \quantTit{UCF101 - DVF} & \quantTit{Xiph - 1K} & \quantTit{Xiph - 2K}
        \vspace{-0.1cm}\\
            & & \quantTit{Baker~\etal~\cite{Baker_IJCV_2011}} & \quantTit{Xue~\etal~\cite{Xue_IJCV_2019}} & \quantTit{Liu~\etal~\cite{Liu_ICCV_2017}} & \quantTit{(4K resized to 1K)} & \quantTit{(4K resized to 2K)}
        \\ \cmidrule(l{2pt}r{2pt}){3-4} \cmidrule(l{2pt}r{2pt}){5-6} \cmidrule(l{2pt}r{2pt}){7-8} \cmidrule(l{2pt}r{2pt}){9-10} \cmidrule(l{2pt}r{2pt}){11-12}
            & {\vspace{-0.29cm} \scriptsize training \linebreak dataset} & \quantSec{PSNR} \linebreak \quantInd{$\uparrow$} & {\vspace{-0.29cm} \scriptsize relative \linebreak improvement} & \quantSec{PSNR} \linebreak \quantInd{$\uparrow$} & {\vspace{-0.29cm} \scriptsize relative \linebreak improvement} & \quantSec{PSNR} \linebreak \quantInd{$\uparrow$} & {\vspace{-0.29cm} \scriptsize relative \linebreak improvement} & \quantSec{PSNR} \linebreak \quantInd{$\uparrow$} & {\vspace{-0.29cm} \scriptsize relative \linebreak improvement} & \quantSec{PSNR} \linebreak \quantInd{$\uparrow$} & {\vspace{-0.29cm} \scriptsize relative \linebreak improvement}
        \\ \midrule
SepConv - $\mathcal{L}_1$ & proprietary & \quantVal{35.73} & \quantVal{-} & \quantVal{33.80} & \quantVal{-} & \quantVal{34.79} & \quantVal{-} & \quantVal{36.22} & \quantVal{-} & \quantVal{34.77} & \quantVal{-}
\\
Ours - $\mathcal{L}_{\textit{Ctx}}$ & Vimeo-90k & \quantVal{37.28} & \quantVal{\text{+ } 1.55 \text{ dB}} & \quantVal{34.83} & \quantVal{\text{+ } 1.03 \text{ dB}} & \quantVal{35.24} & \quantVal{\text{+ } 0.45 \text{ dB}} & \quantVal{36.83} & \quantVal{\text{+ } 0.61 \text{ dB}} & \quantVal{34.84} & \quantVal{\text{+ } 0.07 \text{ dB}}
\\
Ours - $\mathcal{L}_{\textit{Ctx}}$ - $8\hspace{-0.03cm}\times$ & --- \raisebox{-0.09cm}{''} --- & \quantFirst{37.46} & \quantVal{\text{+ } 0.18 \text{ dB}} & \quantFirst{34.97} & \quantVal{\text{+ } 0.14 \text{ dB}} & \quantFirst{35.29} & \quantVal{\text{+ } 0.05 \text{ dB}} & \quantFirst{37.00} & \quantVal{\text{+ } 0.17 \text{ dB}} & \quantFirst{35.10} & \quantVal{\text{+ } 0.26 \text{ dB}}
        \\ \bottomrule
    \end{tabularx}\vspace{-0.2cm}
    \captionof{table}{Quantitative comparison with SepConv. We list two separate results of our proposed approach, one without and one with self-ensembling. The self-ensembling is denoted by $8\hspace{-0.03cm}\times$ as it represents a combination of eight independent estimates.}\vspace{-0.4cm}
    \label{tbl:sepconv}
\end{figure*}

\subsection{Ablation Experiments}
\label{sec:ablation}

We analyze how our proposed techniques affect the interpolation quality in Table~\ref{tbl:ablation}. In short, each technique improves the synthesis quality in terms of PSNR across a variety of datasets. The results on the UCF101 samples as well as the 2K version of the Xiph videos are relatively inconsistent though. However, this behavior is not surprising as several of the UCF101 samples are invalid where the ground truth is identical to either the first or the second input frame (like for examples 1, 141, or 271). As for the high-resolution Xiph videos, the amount of inter-frame motion that is present in in-the-wild 2K footage is expected to exceed the maximum magnitude of $51$ pixels that our adaptive separable kernels can compensate for. We also note that our reimplementation is subject to worse results than the original SepConv~\cite{Niklaus_ICCV_2017} on all datasets except the test split of our training dataset. This finding indicates that there are better training datasets than Vimeo-90k~\cite{Xue_IJCV_2019} for supervising video frame interpolation tasks and that our proposed SepConv\raisebox{0.2ex}{\footnotesize++} could perform even better if it had been supervised on the dataset from~\cite{Niklaus_ICCV_2017}.

\subsection{Self-ensembling for Frame Interpolation}
\label{sec:ensembling}

Our findings on self-ensembling for video frame interpolation summarized in Table~\ref{tbl:ensemble} where we take the mean as well as the median of up to sixteen independent prediction. These findings indicate that any form of self-ensembling is superior to a singular prediction, while taking the mean or taking the median is similarly effective. However, there are diminishing returns in the number of predictions.

Next, we analyze the effect of self-ensembling on all methods that we compare to in this paper. Specifically, we take the mean of eight independent predictions since using sixteen predictions takes much more compute while providing little benefit. As shown in Table~\ref{tbl:ensothers}, all methods benefit from self-ensembling across all datasets. However, self-ensembling has little benefit for practical applications of these frame interpolation techniques since they can already take minutes to process a single second of high-resolution footage. By combining eight independent predictions, this processing time can now become tens of minutes to process a single second of high-resolution footage which is beyond the threshold of being practical for many applications.

\subsection{Comparison with SepConv}

The premise of our paper is that an older and simpler frame interpolation approach, namely SepConv~\cite{Niklaus_ICCV_2017}, can be optimized to produce near state-of-the-art results. In this section, we compare our SepConv\raisebox{0.2ex}{\footnotesize++} with the original SepConv. Specifically, we show a representative qualitative result in Figure~\ref{fig:qualitative} which demonstrates the efficacy of our proposed techniques. Please also consider our supplementary video to better examine this as well as additional examples in motion. Our quantitative comparison in Table~\ref{tbl:sepconv} further shows that our proposed techniques are effective across a variety of datasets as long as the inter-frame motion does not exceed the kernel size (as it occurs for the 2K version of Xiph). Note that this table lists our results with and without self-ensembling for fairness since self-ensembling can easily be applied to all video frame interpolation methods.

\begin{figure*}\centering
    \setlength{\tabcolsep}{0.0cm}
    \renewcommand{\arraystretch}{1.2}
    \newcommand{\quantTit}[1]{\multicolumn{2}{c}{\scriptsize #1}}
    \newcommand{\quantSec}[1]{\scriptsize #1}
    \newcommand{\quantInd}[1]{\scriptsize #1}
    \newcommand{\quantVal}[1]{\scalebox{0.83}[1.0]{$ #1 $}}
    \newcommand{\quantFirst}[1]{\usolid{\scalebox{0.83}[1.0]{$ #1 $}}}
    \newcommand{\quantSecond}[1]{\udotted{\scalebox{0.83}[1.0]{$ #1 $}}}
    \footnotesize
    \begin{tabularx}{\textwidth}{@{\hspace{0.1cm}} X P{1.9cm} P{1.08cm} @{\hspace{-0.31cm}} P{1.85cm} P{1.08cm} @{\hspace{-0.31cm}} P{1.85cm} P{1.08cm} @{\hspace{-0.31cm}} P{1.85cm} P{1.08cm} @{\hspace{-0.31cm}} P{1.85cm} P{1.08cm} @{\hspace{-0.31cm}} P{1.85cm}}
        \toprule
            & & \quantTit{Middlebury} & \quantTit{Vimeo-90k} & \quantTit{UCF101 - DVF} & \quantTit{Xiph - 1K} & \quantTit{Xiph - 2K}
        \vspace{-0.1cm}\\
            & & \quantTit{Baker~\etal~\cite{Baker_IJCV_2011}} & \quantTit{Xue~\etal~\cite{Xue_IJCV_2019}} & \quantTit{Liu~\etal~\cite{Liu_ICCV_2017}} & \quantTit{(4K resized to 1K)} & \quantTit{(4K resized to 2K)}
        \\ \cmidrule(l{2pt}r{2pt}){3-4} \cmidrule(l{2pt}r{2pt}){5-6} \cmidrule(l{2pt}r{2pt}){7-8} \cmidrule(l{2pt}r{2pt}){9-10} \cmidrule(l{2pt}r{2pt}){11-12}
            & {\vspace{0.04cm} \scriptsize venue} & \quantSec{PSNR} \linebreak \quantInd{$\uparrow$} & {\vspace{-0.29cm} \scriptsize absolute \linebreak rank} & \quantSec{PSNR} \linebreak \quantInd{$\uparrow$} & {\vspace{-0.29cm} \scriptsize absolute \linebreak rank} & \quantSec{PSNR} \linebreak \quantInd{$\uparrow$} & {\vspace{-0.29cm} \scriptsize absolute \linebreak rank} & \quantSec{PSNR} \linebreak \quantInd{$\uparrow$} & {\vspace{-0.29cm} \scriptsize absolute \linebreak rank} & \quantSec{PSNR} \linebreak \quantInd{$\uparrow$} & {\vspace{-0.29cm} \scriptsize absolute \linebreak rank}
        \\ \midrule
SepConv - $\mathcal{L}_1$ & ICCV 2017 & \quantVal{35.73} & \quantVal{\hphantom{0}8^\text{\parbox{0.15cm}{th}} \text{ of } 10} & \quantVal{33.80} & \quantVal{10^\text{\parbox{0.15cm}{th}} \text{ of } 10} & \quantVal{34.79} & \quantVal{\hphantom{0}9^\text{\parbox{0.15cm}{th}} \text{ of } 10} & \quantVal{36.22} & \quantVal{\hphantom{0}9^\text{\parbox{0.15cm}{th}} \text{ of } 10} & \quantVal{34.77} & \quantVal{\hphantom{0}8^\text{\parbox{0.15cm}{th}} \text{ of } 10}
\\
CtxSyn - $\mathcal{L}_{\textit{Lap}}$ & CVPR 2018 & \quantVal{36.93} & \quantVal{\hphantom{0}4^\text{\parbox{0.15cm}{th}} \text{ of } 10} & \quantVal{34.39} & \quantVal{\hphantom{0}8^\text{\parbox{0.15cm}{th}} \text{ of } 10} & \quantVal{34.62} & \quantVal{10^\text{\parbox{0.15cm}{th}} \text{ of } 10} & \quantVal{36.87} & \quantVal{\hphantom{0}3^\text{\parbox{0.15cm}{rd}} \text{ of } 10} & \quantVal{35.72} & \quantVal{\hphantom{0}3^\text{\parbox{0.15cm}{rd}} \text{ of } 10}
\\
DAIN & CVPR 2019 & \quantVal{36.69} & \quantVal{\hphantom{0}7^\text{\parbox{0.15cm}{th}} \text{ of } 10} & \quantVal{34.70} & \quantVal{\hphantom{0}6^\text{\parbox{0.15cm}{th}} \text{ of } 10} & \quantVal{35.00} & \quantVal{\hphantom{0}7^\text{\parbox{0.15cm}{th}} \text{ of } 10} & \quantVal{36.78} & \quantVal{\hphantom{0}5^\text{\parbox{0.15cm}{th}} \text{ of } 10} & \quantSecond{35.93} & \quantVal{\hphantom{0}2^\text{\parbox{0.15cm}{nd}} \text{ of } 10}
\\
CAIN & AAAI 2020 & \quantVal{35.11} & \quantVal{10^\text{\parbox{0.15cm}{th}} \text{ of } 10} & \quantVal{34.65} & \quantVal{\hphantom{0}7^\text{\parbox{0.15cm}{th}} \text{ of } 10} & \quantVal{34.98} & \quantVal{\hphantom{0}8^\text{\parbox{0.15cm}{th}} \text{ of } 10} & \quantVal{36.21} & \quantVal{10^\text{\parbox{0.15cm}{th}} \text{ of } 10} & \quantVal{35.18} & \quantVal{\hphantom{0}4^\text{\parbox{0.15cm}{th}} \text{ of } 10}
\\
EDSC - $\mathcal{L}_C$ & arXiv 2020 & \quantVal{36.82} & \quantVal{\hphantom{0}5^\text{\parbox{0.15cm}{th}} \text{ of } 10} & \quantVal{34.83} & \quantVal{\hphantom{0}4^\text{\parbox{0.15cm}{th}} \text{ of } 10} & \quantVal{35.13} & \quantVal{\hphantom{0}6^\text{\parbox{0.15cm}{th}} \text{ of } 10} & \quantVal{36.73} & \quantVal{\hphantom{0}6^\text{\parbox{0.15cm}{th}} \text{ of } 10} & \quantVal{\text{OOM}} & \quantVal{\text{OOM}}
\\
AdaCoF & CVPR 2020 & \quantVal{35.72} & \quantVal{\hphantom{0}9^\text{\parbox{0.15cm}{th}} \text{ of } 10} & \quantVal{34.35} & \quantVal{\hphantom{0}9^\text{\parbox{0.15cm}{th}} \text{ of } 10} & \quantVal{35.16} & \quantVal{\hphantom{0}4^\text{\parbox{0.15cm}{th}} \text{ of } 10} & \quantVal{36.26} & \quantVal{\hphantom{0}8^\text{\parbox{0.15cm}{th}} \text{ of } 10} & \quantVal{34.82} & \quantVal{\hphantom{0}7^\text{\parbox{0.15cm}{th}} \text{ of } 10}
\\
SoftSplat - $\mathcal{L}_{\textit{Lap}}$ & CVPR 2020 & \quantFirst{38.42} & \quantVal{\hphantom{0}1^\text{\parbox{0.15cm}{st}} \text{ of } 10} & \quantFirst{36.10} & \quantVal{\hphantom{0}1^\text{\parbox{0.15cm}{st}} \text{ of } 10} & \quantFirst{35.39} & \quantVal{\hphantom{0}1^\text{\parbox{0.15cm}{st}} \text{ of } 10} & \quantFirst{37.96} & \quantVal{\hphantom{0}1^\text{\parbox{0.15cm}{st}} \text{ of } 10} & \quantFirst{36.63} & \quantVal{\hphantom{0}1^\text{\parbox{0.15cm}{st}} \text{ of } 10}
\\
BMBC & ECCV 2020 & \quantVal{36.79} & \quantVal{\hphantom{0}6^\text{\parbox{0.15cm}{th}} \text{ of } 10} & \quantSecond{35.06} & \quantVal{\hphantom{0}2^\text{\parbox{0.15cm}{nd}} \text{ of } 10} & \quantVal{35.16} & \quantVal{\hphantom{0}4^\text{\parbox{0.15cm}{th}} \text{ of } 10} & \quantVal{36.59} & \quantVal{\hphantom{0}7^\text{\parbox{0.15cm}{th}} \text{ of } 10} & \quantVal{\text{OOM}} & \quantVal{\text{OOM}}
\\
Ours - $\mathcal{L}_{\textit{Ctx}}$ & N/A & \quantVal{37.28} & \quantVal{\hphantom{0}3^\text{\parbox{0.15cm}{rd}} \text{ of } 10} & \quantVal{34.83} & \quantVal{\hphantom{0}4^\text{\parbox{0.15cm}{th}} \text{ of } 10} & \quantVal{35.24} & \quantVal{\hphantom{0}3^\text{\parbox{0.15cm}{rd}} \text{ of } 10} & \quantVal{36.83} & \quantVal{\hphantom{0}4^\text{\parbox{0.15cm}{th}} \text{ of } 10} & \quantVal{34.84} & \quantVal{\hphantom{0}6^\text{\parbox{0.15cm}{th}} \text{ of } 10}
\\
Ours - $\mathcal{L}_{\textit{Ctx}}$ - $8\hspace{-0.03cm}\times$ & --- \raisebox{-0.09cm}{''} --- & \quantSecond{37.46} & \quantVal{\hphantom{0}2^\text{\parbox{0.15cm}{nd}} \text{ of } 10} & \quantVal{34.97} & \quantVal{\hphantom{0}3^\text{\parbox{0.15cm}{rd}} \text{ of } 10} & \quantSecond{35.29} & \quantVal{\hphantom{0}2^\text{\parbox{0.15cm}{nd}} \text{ of } 10} & \quantSecond{37.00} & \quantVal{\hphantom{0}2^\text{\parbox{0.15cm}{nd}} \text{ of } 10} & \quantVal{35.10} & \quantVal{\hphantom{0}5^\text{\parbox{0.15cm}{th}} \text{ of } 10}
        \\ \bottomrule
    \end{tabularx}\vspace{-0.2cm}
    \captionof{table}{Quantitative comparison with recent approaches for video frame interpolation. In addition to highlighting the best result by underlining it, we emphasize the second-best result via a dotted underline. Note that some methods were unable to run on 2K footage due to exceeding the $16$ gigabytes of memory available on our graphics card (denoted as ``OOM'').}\vspace{-0.4cm}
    \label{tbl:quantitative}
\end{figure*}

\begin{figure}\centering
    \setlength{\tabcolsep}{0.0cm}
    \renewcommand{\arraystretch}{1.2}
    \newcommand{\quantTit}[1]{\multicolumn{2}{c}{\scriptsize #1}}
    \newcommand{\quantSec}[1]{\scriptsize #1}
    \newcommand{\quantInd}[1]{\scriptsize #1}
    \newcommand{\quantVal}[1]{\scalebox{0.83}[1.0]{$ #1 $}}
    \newcommand{\quantFirst}[1]{\usolid{\scalebox{0.83}[1.0]{$ #1 $}}}
    \newcommand{\quantSecond}[1]{\udotted{\scalebox{0.83}[1.0]{$ #1 $}}}
    \footnotesize
    \begin{tabularx}{\columnwidth}{@{\hspace{0.1cm}} X P{1.08cm} @{\hspace{-0.31cm}} P{1.85cm} P{0.2cm} P{1.08cm} @{\hspace{-0.31cm}} P{1.85cm}}
        \toprule
            & \quantTit{Middlebury} && \quantTit{Middlebury}
        \vspace{-0.1cm}\\
            & \quantTit{(mean error)} && \quantTit{(median error)}
        \\ \cmidrule(l{2pt}r{2pt}){2-3} \cmidrule(l{2pt}r{2pt}){5-6}
            & \quantSec{IE} \linebreak \quantInd{$\downarrow$} & {\vspace{-0.29cm} \scriptsize absolute \linebreak rank} && \quantSec{IE} \linebreak \quantInd{$\downarrow$} & {\vspace{-0.29cm} \scriptsize absolute \linebreak rank}
        \\ \midrule
SepConv - $\mathcal{L}_1$ & \quantVal{5.61} & \quantVal{8^\text{\parbox{0.15cm}{th}} \text{ of } 9} && \quantVal{5.44} & \quantVal{8^\text{\parbox{0.15cm}{th}} \text{ of } 9}
\\
CtxSyn - $\mathcal{L}_{\textit{Lap}}$ & \quantVal{5.28} & \quantVal{7^\text{\parbox{0.15cm}{th}} \text{ of } 9} && \quantVal{4.77} & \quantVal{7^\text{\parbox{0.15cm}{th}} \text{ of } 9}
\\
DAIN & \quantVal{4.86} & \quantVal{6^\text{\parbox{0.15cm}{th}} \text{ of } 9} && \quantVal{4.69} & \quantVal{5^\text{\parbox{0.15cm}{th}} \text{ of } 9}
\\
CAIN & \quantVal{-} & \quantVal{-} && \quantVal{-} & \quantVal{-}
\\
EDSC - $\mathcal{L}_C$ & \quantVal{4.72} & \quantVal{4^\text{\parbox{0.15cm}{th}} \text{ of } 9} && \quantVal{4.69} & \quantVal{5^\text{\parbox{0.15cm}{th}} \text{ of } 9}
\\
AdaCoF & \quantVal{4.75} & \quantVal{5^\text{\parbox{0.15cm}{th}} \text{ of } 9} && \quantVal{4.48} & \quantVal{4^\text{\parbox{0.15cm}{th}} \text{ of } 9}
\\
SoftSplat - $\mathcal{L}_{\textit{Lap}}$ & \quantFirst{4.22} & \quantVal{1^\text{\parbox{0.15cm}{st}} \text{ of } 9} && \quantFirst{3.97} & \quantVal{1^\text{\parbox{0.15cm}{st}} \text{ of } 9}
\\
BMBC & \quantVal{4.48} & \quantVal{3^\text{\parbox{0.15cm}{rd}} \text{ of } 9} && \quantVal{4.16} & \quantVal{3^\text{\parbox{0.15cm}{rd}} \text{ of } 9}
\\
Ours - $\mathcal{L}_{\textit{Ctx}}$ & \quantSecond{4.45} & \quantVal{2^\text{\parbox{0.15cm}{nd}} \text{ of } 9} && \quantSecond{4.13} & \quantVal{2^\text{\parbox{0.15cm}{nd}} \text{ of } 9}
        \\ \bottomrule
    \end{tabularx}\vspace{-0.2cm}
    \captionof{table}{Quantitative results on the official Middlebury benchmark~\cite{Baker_IJCV_2011}. This benchmark does not list CAIN~\cite{Choi_AAAI_2020}.}\vspace{-0.4cm}
    \label{tbl:middlebury}
\end{figure}

\subsection{Comparison with Others}

Even though we base our approach off of an older and simpler frame interpolation technique, we are able to achieve near state-of-the-art quality. To demonstrate this, we compare our SepConv\raisebox{0.2ex}{\footnotesize++} to competitive approaches for frame interpolation based on kernel prediction (SepConv~\cite{Niklaus_ICCV_2017}, EDSC~\cite{Cheng_ARXIV_2020}, and AdaCoF~\cite{Lee_CVPR_2020}), based on optical flow estimation and compensation (CtxSyn~\cite{Niklaus_CVPR_2018}, DAIN~\cite{Bao_CVPR_2019}, SoftSplat~\cite{Niklaus_CVPR_2020}, and BMBC~\cite{Park_ECCV_2020}), and based on directly synthesizing the intermediate frame (CAIN~\cite{Choi_AAAI_2020}). We summarize the findings in Table~\ref{tbl:quantitative}, where we separately list our results with and without self-ensembling for fairness. In summary, our proposed approach is only outperformed by SoftSplat and generally does not fair well on the 2K version of the Xiph footage where the inter-frame motion exceeds what our adaptive separable kernels with $51$ pixels can compensate for. However, SoftSplat was additionally supervised on training data with ground truth optical flow, whereas our approach was solely supervised on the Vimeo-90k~\cite{Xue_IJCV_2019} dataset.

We also submitted our results to the organizers of the Middlebury benchmark~\cite{Baker_IJCV_2011} where our SepConv\raisebox{0.2ex}{\footnotesize++} ranks second in terms of interpolation error among all published methods (currently not publicly visible, please see our supplementary material). We show a summary of the benchmark in Table~\ref{tbl:middlebury} with all methods that we compare to in this paper.

\subsection{Discussion}

While we were able to show that it is possible to achieve near state-of-the-art video frame interpolation results using an older technique by carefully optimizing it, we did not fundamentally alter its image formation model. As such, the two main limitations of video frame interpolation via adaptive separable convolutions remain. First, they often do not produce satisfying interpolation results on high-resolution input footage due to the limited kernel size. This is exemplified by the relatively poor performance of SepConv\raisebox{0.2ex}{\footnotesize++} on the 2K version of the Xiph footage. Second, they are limited to synthesizing the interpolation result at the temporal position that they have been supervised on. To synthesize interpolation results at $t=0.75$ instead of $t=0.5$, one would either have to train a kernel prediction network with the ground truth at $t=0.75$ or first synthesize the interpolation result at $t=0.5$ and then recursively use this result as well as the input at $t=1$ to yield $t=0.75$. This limits practical applications where, for example, a given video with $50$ frames per second needs to be converted to $60$ frames per second.

\section{Conclusion}
\label{sec:conclusion}

In this paper, we show, somewhat surprisingly, that it is possible to achieve near state-of-the-art video frame interpolation results with an older, simpler approach by carefully optimizing it. Our optimizations are conceptually intuitive, effective in improving the interpolation quality, and are directly applicable to a variety of frame interpolation techniques. Furthermore, while our paper focuses on improving adaptive separable convolutions for the purpose of frame interpolation, some of our proposed techniques may be applicable to related applications as well, such as image denoising, joint image filtering, or video prediction.

{\small
\bibliographystyle{ieee_fullname}
\bibliography{egbib} 
}

\end{document}